\documentclass[conference]{IEEEtran}
\usepackage{times}

\usepackage[numbers]{natbib}
\usepackage{multicol}
\usepackage[bookmarks=true]{hyperref}

\usepackage[linesnumbered,ruled,vlined]{algorithm2e}
\usepackage{amsmath} 
\usepackage{amsfonts}
\usepackage{graphicx}
\usepackage{xcolor}

\usepackage{booktabs, multirow} 
\usepackage{soul}
\usepackage{xcolor,colortbl} 
\usepackage{changepage,threeparttable} 

\begin{document}

\title{Latent Diffeomorphic Co-Design of End-Effectors for Deformable and Fragile Object Manipulation}




%
\author{\authorblockN{Kei Ikemura*,
Yifei Dong*,
Florian T. Pokorny}
}

\maketitle
\def\thefootnote{*}\footnotetext{Equal contributions.}

\begin{abstract}
Manipulating deformable and fragile objects remains a fundamental challenge in robotics due to complex contact dynamics and strict requirements on object integrity. Existing approaches typically optimize either end-effector design or control strategies in isolation, limiting achievable performance. In this work, we present the first co-design framework that jointly optimizes end-effector morphology and manipulation control for deformable and fragile object manipulation. We introduce (1) a latent diffeomorphic shape parameterization enabling expressive yet tractable end-effector geometry optimization, (2) a stress-aware bi-level co-design pipeline coupling morphology and control optimization, and (3) a privileged-to-pointcloud policy distillation scheme for zero-shot real-world deployment. We evaluate our approach on challenging food manipulation tasks, including grasping and pushing jelly and scooping fillets. Simulation and real-world experiments demonstrate the effectiveness of the proposed method.
\def\thefootnote{1}\footnotetext{
Supplementary video: {\url{https://youtu.be/LDAGx_XiTP4}}.
Funded by the European Commission under the Horizon Europe Framework Program project SoftEnable, grant number 101070600, {\url{https://softenable.eu/}}.
Contact: {\tt\small \{ikemura, yifeid\}@kth.se}.
}
\end{abstract}

\IEEEpeerreviewmaketitle


\section{Introduction}
Deformable and Fragile Object Manipulation (DFOM) is central to many real-world robotic applications, including food handling~\cite{wang2022challenges, swann2025dexfruit}, surgical assistance~\cite{masui2024vision, wang2025image}, and caregiving~\cite{wang2023deformable}. Despite recent progress, DFOM remains challenging due to several fundamental factors: (i) deformable objects exhibit high-dimensional states and complex, nonlinear contact dynamics, (ii) large variations in physical properties such as stiffness, fracture thresholds, and surface friction hinder generalization across objects and environments, and (iii) gentle manipulation requires precise, reactive control to preserve object integrity throughout dynamic interaction.

Most existing DFOM approaches address these challenges from two complementary directions: end-effector hardware design or control strategy development. On the hardware side, researchers design rigid~\cite{allison2024hashi}, soft~\cite{enomoto2023delicate, wang2020dual}, or hybrid grippers~\cite{gafer2020quad} tailored for fragile object handling. While such designs often outperform generic parallel-jaw grippers, they are typically handcrafted by domain experts and rarely optimized directly against empirical task performance, with control strategies remaining relatively simple. On the control side, methods assume fixed, standard grippers and focus on advanced control using additional sensing modalities to enable gentle interaction~\cite{cuiral2023contour, gong2024deformability, langsfeld2016robotic, wen2020force, yagawa2023learning}. However, this separation limits the achievable performance, as the controller cannot exploit morphology variations that could fundamentally improve contact geometry and force distribution.

\begin{figure}[t]
\centering
\includegraphics[width=0.99\linewidth]{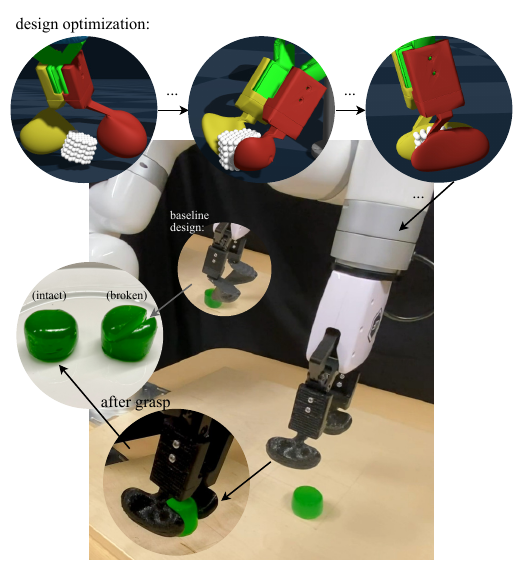}
\caption{We jointly optimize end-effector morphology and motion-adaptive control to enable safe, gentle manipulation of deformable and fragile objects. The co-designed end-effector reshapes contact geometry and force distribution, achieving reliable grasping while preserving object integrity, where baseline designs fail and cause breakage. 
}
\label{fig:teaser}
\end{figure}

Beyond treating hardware and control independently, joint optimization of robot morphology and control, commonly referred to as co-design, offers a principled mechanism to exploit their mutual dependence. In primate evolution, neural and bodily adaptations are interconnected~\cite{baker2025human}. For instance, the co-evolution of human hand morphology (e.g., longer and opposable thumbs) and motor control has enabled precise and dexterous manipulation of delicate materials. Similar principles suggest that co-design can substantially enhance robotic manipulation capabilities~\cite{wang2025embodied}. Despite its potential, co-design has remained largely unexplored for deformable and fragile object manipulation, with most existing co-design studies focusing on rigid-body tasks~\cite{gao2025vlmgineer, guo2024learning, xu2021end}. This gap motivates our work: simultaneously optimizing end-effector morphology and manipulation strategy to address the unique challenges posed by DFOM.

To this end, we present the first co-design framework explicitly targeting deformable and fragile object manipulation. 
Our contributions are threefold: (i) a novel shape parameterization for co-design, Latent DiffeoMorphism (LDM), that enables expressive yet physically consistent end-effector design while remaining amenable to gradient-free optimization, (ii) a bi-level co-design pipeline for DFOM that leverages privileged simulator feedback to jointly optimize morphology and control strategies, and (iii) a sim-to-real transfer strategy based on teacher–student policy distillation that enables zero-shot deployment on real robotic systems.

Building on these components, we develop a unified framework that integrates an expressive diffeomorphic design space, and a bi-level optimization procedure followed by teacher-student policy distillation. We evaluate the proposed method in both simulation and real-world experiments on challenging food manipulation tasks, such as grasping and pushing jelly. Results demonstrate that the co-designed end-effectors and design-conditioned control strategies achieve reliable task execution while substantially reducing contact-induced damage compared to baselines.

\begin{figure*}[t]
\centering
\includegraphics[width=0.865\linewidth]{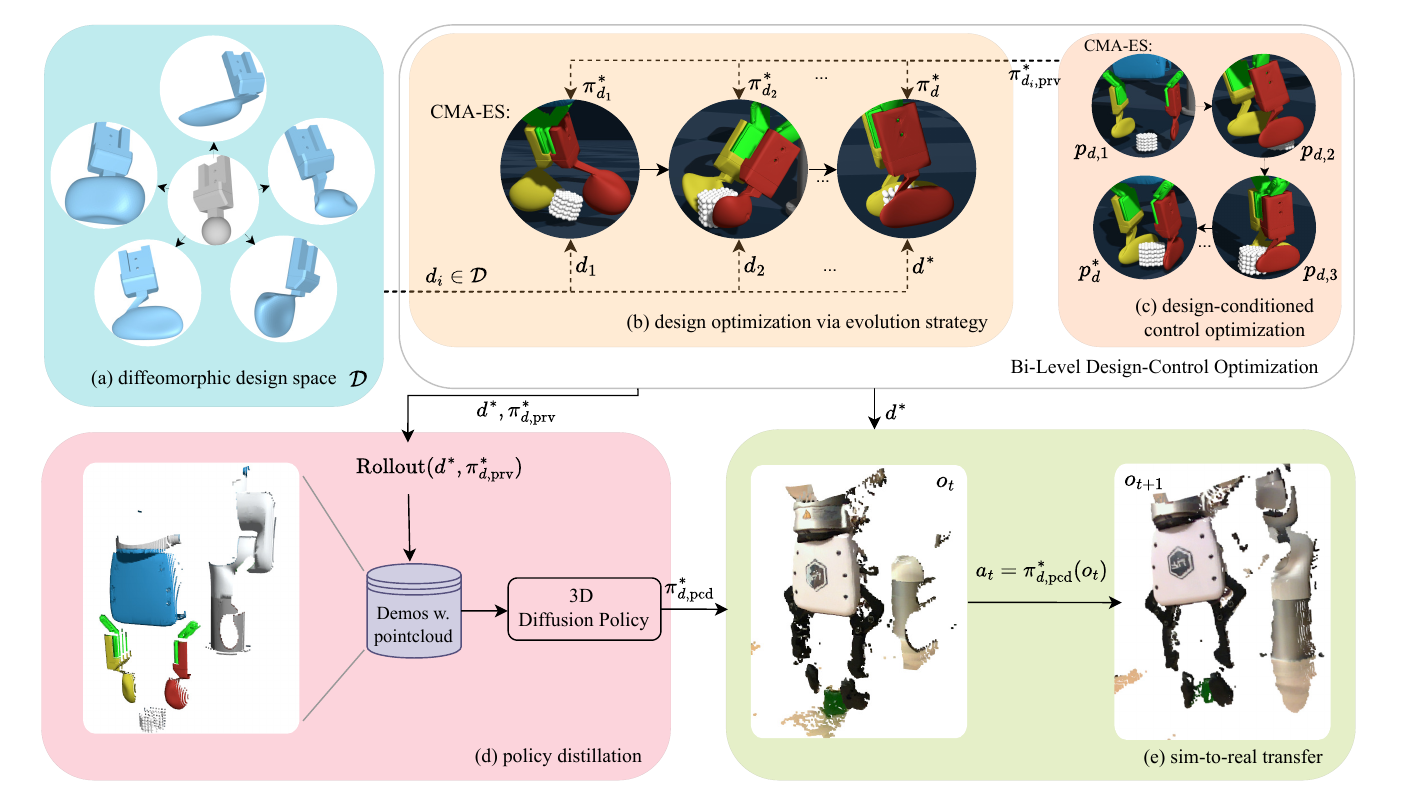}
\caption{Overview of the proposed co-design framework for deformable and fragile object manipulation. 
(a) A diffeomorphic design space $\mathcal{D}$ parameterizes physically valid end-effector geometries. 
(b) A derivative-free evolutionary optimizer (CMA-ES) searches $\mathcal{D}$ to identify an optimal design $d^*$ that maximizes task performance while minimizing contact-induced damage. 
(c) For each sampled design $d_i$, a design-conditioned controller $\pi^*_{d_i,\text{prv}}$ is optimized using simulator's privileged signals such as object stress. 
(d) Rollouts of the privileged policy on the optimized design~$d$ are used to collect pointcloud observations and train a student diffusion policy $\pi^*_{d,\text{pcd}}$. 
(e) The distilled policy is deployed zero-shot on the real robot using pointcloud observations, preserving gentle manipulation behaviors learned in simulation.
}
\label{fig:overview}
\end{figure*}


\section{Related Work}
\subsection{Deformable and Fragile Object Manipulation}
Research on deformable and fragile object manipulation has largely pursued object integrity through (i) specialized end-effector hardware, (ii) additional sensing for feedback control, or (iii) learned policies. Hardware-centric approaches commonly employ soft or compliant fingertips/end-effectors to reduce peak contact forces and distribute pressure~\cite{maruyama2013delicate,wang2021circular}, but these designs typically demand substantial manual engineering and are often difficult to adapt across tasks and object categories. Sensing-centric approaches instead rely on tactile or force measurements to regulate interaction and avoid damage: Huang et al.~\cite{huang2019learning} learn gentle manipulation with fingertip touch sensing, Yagawa et al.~\cite{yagawa2023learning} anticipate fracture using tactile feedback, and Lee et al.~\cite{lee2025grasping} study deformable grasping in visuo-tactile simulation. While effective, such methods introduce extra hardware, calibration burden, and cost, which can be undesirable in practical deployments.

To reduce reliance on physical sensors, another line of work estimates interaction forces from vision. Wang et al.~\cite{wang2025image} infer contact forces using structured light, Jung et al.~\cite{jung2020vision} estimate suture tension from video and tool poses, and Masui et al.~\cite{masui2024vision} detect unsafe force events from surgical imagery. In contrast to sensor-heavy pipelines or perception-based force inference, our method leverages privileged stress signals available in modern soft-body simulators as a physical supervision source. This enables (i) {direct optimization} of end-effector morphology and manipulation strategy against an explicit object integrity objective, and (ii) {zero-shot real-world deployment} via domain randomization and policy distillation, without requiring on-finger tactile/force sensors or real-world training.  


\subsection{Robot Co-Design}
Robot co-design problems are commonly formulated as bi-level optimization, with a lower-level control optimization and an upper-level design optimization. Most prior works treat these two levels separately. At the control level, learning design-conditioned policies using reinforcement learning is a common approach~\cite{ringel2025text2robot, dong2025cagecoopt, belmonte2022meta}, but the large dimensionality of design spaces make it difficult to learn a general-purpose controller that generalizes across embodiments. Alternatively, heuristic or model-based controllers have been adopted for the lower-level optimization~\cite{kim2021mo}. Compared with prior approaches that rely on hand-scripted motion sequences, our method optimizes motion-adaptive control strategies that explicitly account for variations in end-effector geometry as well as object shape and pose.
At the design level, given a design-conditioned controller, prior works typically search for optimal designs using Bayesian Optimization~\cite{bjelonic2023learning, mockus2005bayesian, pan2021emergent}, Genetic Algorithms~\cite{ha2019reinforcement, sathuluri2023robust}, or neural surrogate models~\cite{ye2025power}. However, these methods are usually limited to low-dimensional and weakly expressive design spaces, which restrict the diversity of achievable designs. In contrast, our approach explicitly targets the trade-off between expressiveness and optimization tractability by introducing a novel diffeomorphic shape parametrization with a compact latent representation. 

Recent works have also formulated co-design as a single Markov Decision Process and learned design and control jointly via end-to-end reinforcement learning~\cite{ikemuraefficient, guo2024learning, yuan2021transform2act}, but these approaches often suffer from severe sample inefficiency in design exploration of high-dimensional design spaces. Co-design driven by vision language models has also recently emerged as a promising direction for leveraging semantic priors and human intent~\cite{gao2025vlmgineer, lin2025robotsmith}. Among the broader co-design literature, our work is the first to explicitly target the unique challenges of deformable and fragile object manipulation~\cite{wang2025embodied}.






\section{Problem Formulation}

We formulate deformable and fragile object manipulation as a co-design problem that jointly optimizes the end-effector geometry and the associated manipulation strategy. Let $d \in \mathcal{D}$ denote a candidate end-effector design in the design space $\mathcal{D}$. 
The co-design problem is formulated as the following bi-level optimization:
\begin{subequations}
\begin{align}
d^\star 
&= \arg\max_{d \in \mathcal{D}} \mathbb{E}_{\xi}\left[ 
J_{\xi}(d, \pi^*_d)
\right], 
\label{eq:bilevel_upper}\\
\text{s.t.}\quad 
\pi^*_d 
&= 
\arg\max_{\pi} \mathbb{E}_{\xi}\left[ 
J_{\xi}(d, \pi)
\right].
\label{eq:bilevel_lower}
\end{align}
\end{subequations}
The lower-level problem~\eqref{eq:bilevel_lower} corresponds to finding a manipulation strategy~$\pi^*_d$ given a specific design $d$. Given the design-conditioned optimal control strategy~$\pi^*_d$, the upper-level problem~\eqref{eq:bilevel_upper} searches for an end-effector design $d$ that yields the best task performance.
The score $J(d, \pi^*_d)$ is an object internal stress-aware task objective that trades off task success and the integrity of fragile objects. 
The expectation is approximated by executing DFOM tasks in randomized task contexts~$\xi$, which incorporate object pose, size and physical properties, robot Cartesian space noises and initial pose, etc.
 
In the following, we first introduce the design space parametrization based on Diffeomorphism~\cite{younes2010shapes, han2024hybrid, sun2022topology}, followed by a motion-adaptive control strategy that leverages privileged simulator information such as soft-body stress and object centroid pose. We then describe the derivative-free design optimization procedure built on the proposed design-conditioned control strategy. Finally, we present a teacher–student distillation framework to train a reactive controller from non-privileged point cloud observations for real-world deployment. The overview of the proposed framework is presented in Fig.~\ref{fig:overview}.



\section{Latent Diffeomorphic Shape Parametrization}

For end-effector co-design, we require a compact yet expressive parametrization of the shape design space $\mathcal{D}$. Existing approaches based on explicit geometry templates~\cite{dong2025cagecoopt, guo2024learning}, voxel grids~\cite{bhatia2021evolution}, or kinematic linkages~\cite{liu2023learning} are often limited in expressiveness and typically restricted to low-dimensional or 2D settings. Given this, we propose \emph{Latent DiffeoMorphism (LDM)}, which combines diffeomorphic shape deformation with data-driven dimensionality reduction to achieve an expressive yet tractable design space.

\begin{figure}[t]
    \centering
    \includegraphics[width=0.6\linewidth]{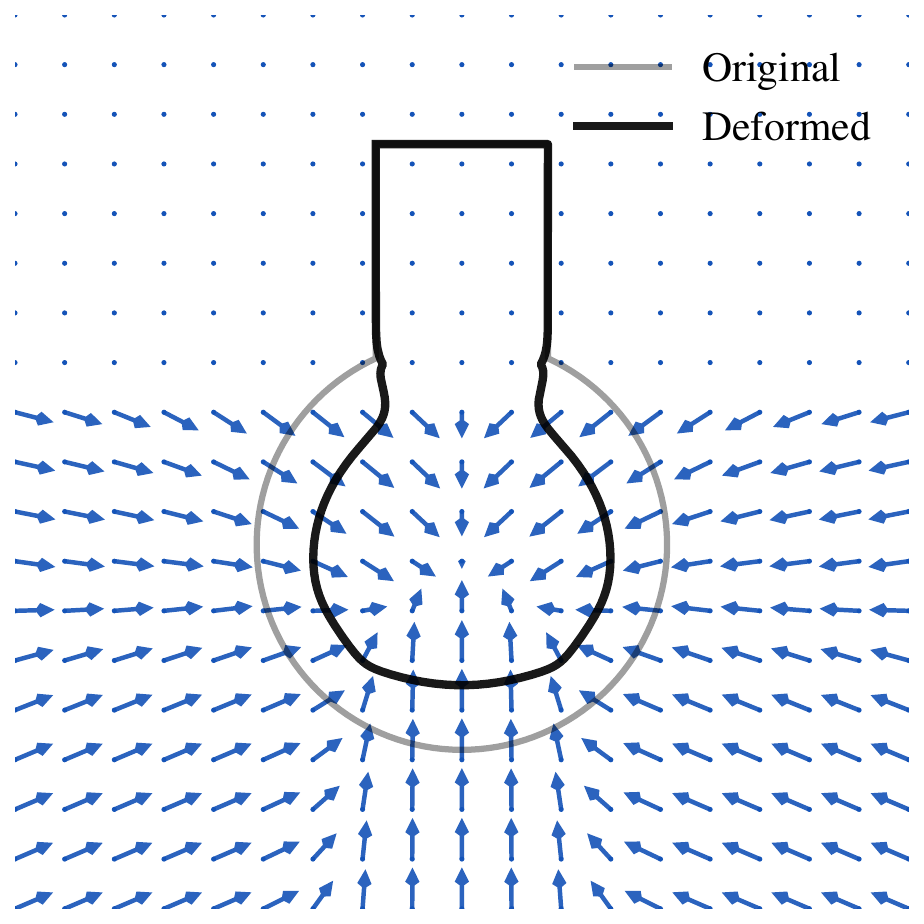}
    \caption{Deformation of a 2D gripper contour under a stationary velocity field (SVF). 
    The upper region is frozen (zero velocity), while the lower region deforms smoothly.}
    \label{fig:svf_gripper}
\end{figure}

\paragraph{Diffeomorphic shape parametrization}
The core idea of LDM is to represent shapes as smooth, invertible deformations of a common base shape. A diffeomorphism is a smooth bijective mapping with a smooth inverse, which guarantees topology preservation and prevents self-penetration during deformation~\cite{younes2010shapes}. Given a base shape mesh, we generate a target shape by integrating a stationary velocity field (SVF) $v:\mathbb{R}^3 \rightarrow \mathbb{R}^3$ over time $t \in [0,1]$, leading to a deformation map $\phi_t$ that satisfies
\begin{equation}
    \frac{d \phi_t(x)}{dt} = v(\phi_t(x)), \quad \phi_0(x) = x.
\end{equation}
Intuitively, each point $x$ on or inside the base mesh is treated as a particle flowing under the velocity field $v$. By integrating this flow from $t=0$ to $t=1$, the point is transported to a new location $\phi_1(x)$, and the collection of all such transported points defines the deformed target shape mesh, as shown in Fig.~\ref{fig:svf_gripper} (an illustration in $\mathbb{R}^2$).
Compared to alternative deformation-based shape parameterization, such as cage-based deformation methods~\cite{ikemuraefficient}, this formulation enforces diffeomorphic mappings and thus avoids self-penetration while retaining high deformation flexibility. Our design space is thus a parameterization of the SVF $v(x)$.

\paragraph{Velocity field parameterization}
We parameterize the stationary velocity field using Radial Basis Functions (RBFs). Specifically, we place $K$ RBF kernels at fixed centers $c_i \in \mathbb{R}^3$ and express the velocity field as
\begin{equation}
    v(x) = \sum_{i=1}^{K} \lambda_i \exp\!\left(-\frac{\|x - c_i\|^2}{2\sigma_i^2}\right),
\end{equation}
where $\lambda_i \in \mathbb{R}^3$ denotes the RBF weight vector and $\sigma_i \in \mathbb{R}$ is the kernel width. Each RBF contributes four scalar parameters, resulting in a $M_\text{raw}$-dimensional deformation parameter vector ($M_\text{raw}=4K$). Empirically, we set $K=18$, which provides sufficient deformation expressiveness, resulting in a $72$-dimensional raw design space.

\begin{figure}[t]
    \centering
    \includegraphics[width=0.75\linewidth]{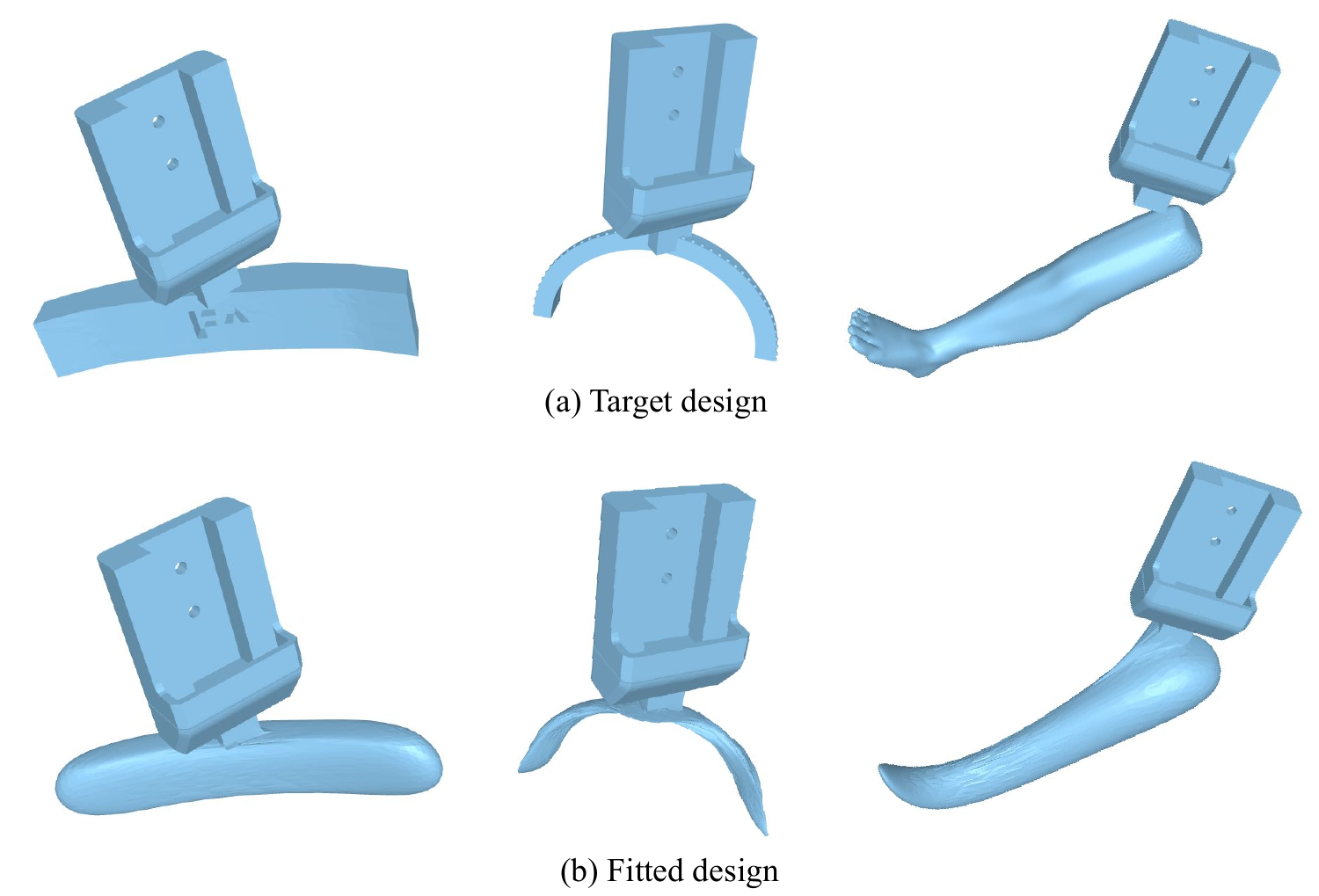}
    \caption{Examples of diffeomorphic deformation parameter fitting. The target shape surface (a) and the fitted mesh (b) are close in terms of Chamfer distance.}
    \label{fig:design-fitting}
\end{figure}

\paragraph{Latent space construction via PCA}
Direct optimization in this high-dimensional space (of $4K$ parameters) is impractical for co-design, where the majority of prior works operate in design spaces with dimensionality below 10~\cite{yang2024evolving, wang2025embodied, belmonte2022meta, he2024morph}.
To address this, we construct a low-dimensional latent design space using data-driven dimensionality reduction. We collect a dataset of $N$ curated finger designs from the Thingi10K dataset~\cite{zhou2016thingi10k}, which is augmented from $N_0$ manual designs ($N_0=150$, $N=1000$ in practice). For each shape, we fit the corresponding diffeomorphic deformation parameters by minimizing the Chamfer distance between the deformed base mesh and the target shape surface with gradient descent (Fig.~\ref{fig:design-fitting}). A dataset thereby is formed in the raw $M_\text{raw}$-dimensional parameter space. We then apply Principal Component Analysis (PCA) to obtain a compact latent representation of dimension $M_{\text{latent}} \ll M_\text{raw}$.

\paragraph{Runtime decoding and shape synthesis}
During co-design optimization, we sample a latent vector $z \in \mathbb{R}^{M_{\text{latent}}}$, map it back to the raw deformation parameter space via the inverse PCA transform, and apply the resulting SVF-based diffeomorphic deformation to the base shape. This leads to a physically valid and topology-consistent candidate design. In this way, LDM enables efficient exploration of a rich, continuous shape space while maintaining a low-dimensional optimization interface suitable for end-effector co-design.

\section{Sim-to-Real End-Effector Co-Design for DFOM}
\subsection{Design-Conditioned Control Strategy}
We address the lower-level problem in Eq.~\eqref{eq:bilevel_lower}: given a design $d \in \mathcal{D}$ parameterized by LDM, compute the corresponding optimal control strategy $\pi_d^*$. We adopt a design-conditioned, motion-adaptive control framework that combines pre-contact confugration optimization with closed-loop motion primitives. Specifically, we first optimize a pre-contact end-effector configuration $p_d^* \in \mathcal{P}$, where the configuration~$p$ includes the end-effector position~$\alpha$, orientation~$R$, and gripper opening width~$w$. A sequence of motion primitives $\pi_d(p_d^*)$ is executed via $p_d^*$ to accomplish the task. For notational simplicity, we omit the explicit dependence on $d$ when the context is clear.

\paragraph{Pre-contact configuration optimization}
The pre-contact configuration is obtained by solving
\begin{equation}
\label{eq:pose_objective}
p^* = \arg\max_{p} f(p; d, \xi)
= \arg\max_{p} \sum_{m \in \mathcal{M}} \lambda_m \, f_m(p; \rho_d, \phi_\xi),
\end{equation}
where $\mathcal{M} = \{\text{pen}, \text{near}\} \cup \mathcal{M}_{\text{task}}$ denotes a set of scoring terms, including penetration avoidance $f_{\text{pen}}$, proximity to the target object $f_{\text{near}}$, and task-specific objectives in $\mathcal{M}_{\text{task}}$. These terms are computed using the signed distance fields (SDFs) of the object $\phi_\xi$ and the end-effector mesh $\rho_d$.

We solve this non-convex optimization problem using Covariance Matrix Adaptation Evolution Strategy (CMA-ES)~\cite{hansen2003reducing}. Although the objective in Eq.~\eqref{eq:pose_objective} is differentiable, SDF- and mesh-based evaluations introduce numerical instability and poor conditioning for gradient-based solvers, making CMA-ES more robust in practice.

To balance computational efficiency and physical fidelity, we adopt a two-stage evaluation strategy that performs large-scale optimization using a rigid-body surrogate and reserves soft-body simulation for a small set of promising candidates. Specifically, thousands of objective evaluations are required per design, making direct soft-body optimization prohibitively expensive. Moreover, because the pre-contact configuration enforces non-penetration constraints, rigid geometric approximations are sufficiently accurate at this stage. We therefore generate $I$ diverse candidate configurations $\{p_i\}_{i=1}^I$ with high objective scores (Algorithm~\ref{alg:codesign}, line~6) using the object's rigid body surrogate, evaluate each candidate in soft-body simulation, and select the best-performing configuration as $p^*$ (lines~7--9).

\paragraph{Adaptive motion primitives}
\label{paragraph:motion-primitive}
After reaching the pre-contact configuration $p$, the robot executes a sequence of motion primitives $\pi(p)$ that adapt online to the task context $\xi$ and simulator feedback. Together, the optimized pre-contact configuration and the subsequent primitives define the full control strategy $\pi$ in Eq.~\eqref{eq:bilevel_lower}. Each primitive is implemented using closed-loop Cartesian controllers, allowing the system to react to variations in end-effector design, object pose and geometry, and contact conditions. Privileged simulator signals, including object stress and centroid motion, are monitored to trigger phase transitions between manipulation stages (e.g., approach-to-contact in grasping).

We adopt this model-based motion-adaptive control strategy instead of reinforcement learning for two main reasons. First, the flexible and relatively high-dimensional design space (around 15-dimensional) induced by LDM makes training a general-purpose design-conditioned RL policy challenging, due to reward design difficulties or severe sample inefficiency under sparse task rewards. 
Second, the proposed controller is interpretable, reliable, and enables efficient generation of large-scale, high-quality demonstrations across diverse end-effector designs without requiring teleoperation. The choice here is validated in Section~\ref{sec:evaluation}, where end-to-end RL for co-design struggles with the high-dimensional design space.



\subsection{End-Effector Design Optimization}
We address the upper-level problem in Eq.~\eqref{eq:bilevel_upper}: given a design-conditioned control strategy $\pi_d^*$, identify the optimal end-effector design $d^*$. For each candidate design $d \in \mathcal{D}$, we evaluate its performance by executing the corresponding control strategy $\pi_d^*(p_d^*)$ in simulation and computing a design score $J(d)$ that trades off task success and object integrity using privileged simulator information. Since this objective is non-differentiable and obtained from stochastic physics rollouts, we solve the resulting black-box optimization problem using a derivative-free optimizer. In the following, we first define the design evaluation score and then describe the optimization procedure.

\paragraph{Design evaluation score}
For each parallel environment $e$ in the physical simulator, the evaluation score of design $d$ is computed as
\begin{equation}
J_e(d)
=
\lambda_{\text{prog}}\, q_{\text{prog}}
+ \lambda_{\text{succ}}\, q_{\text{succ}}
- \lambda_{\bar{\sigma}}\, \bar{\sigma}
- \lambda_{\mathrm{max}}\, \sigma^{\mathrm{max}},
\label{eq:single_objective}
\end{equation}
where $\lambda_{\cdot} > 0$ are weighting coefficients. We use two task success metrics: (i) a continuous metric $q_{\text{prog}} \in \mathbb{R}$ that measures task progress (e.g., object lift distance in grasping), and (ii) a binary success metric $q_{\text{succ}} \in \{0,1\}$ that indicates whether the task is achieved and maintained during the terminal phase of the rollout.

To compute the stress-based fragile object integrity metrics, let $V$ denote the set of soft-body mesh vertices, and let $\sigma_{k,v}$ represent the stress at vertex $v \in V$ and timestep $k \in \{1,\dots,K\}$. We define two complementary stress metrics:
\begin{align}
\bar{\sigma}_{k} &= \frac{1}{|V|}\sum_{v \in V} \sigma_{k,v}, \\
\sigma^{\mathrm{max}}_{k} &= P_s\!\left(\{\sigma_{k,v}\}_{v \in V}\right), \label{eq:max-stress-spatial}
\end{align}
where $P_s(\cdot)$ denotes the top $s\%$ percentile of a set sorted in descending order, indicating $s\%$ of elements in the set are larger than $P_s(\cdot)$. In practice, we use $s=2.5$ and $s=0.0$. The mean stress $\bar{\sigma}_k$ penalizes globally excessive deformation, while $\sigma^{\mathrm{max}}_k$ captures localized stress concentration on a small subset of vertices.

We then aggregate each stress metric temporally over the rollout horizon $K$:
\begin{equation}
\bar{\sigma} = P_s\!\left(\{\bar{\sigma}_{k}\}_{k=1}^{K}\right), 
\quad
\sigma^{\mathrm{max}} = P_s\!\left(\{\sigma^{\mathrm{max}}_{k}\}_{k=1}^{K}\right).
\label{eq:stress-temporal}
\end{equation}
Finally, the overall design evaluation score $J(d)$ is obtained by averaging $J_e(d)$ across all environments (Algorithm~\ref{alg:codesign}, line~11).

\paragraph{Derivative-free design optimization}
We employ CMA-ES as the derivative-free design optimizer $\mathcal{O}_d$, which is well suited for high-dimensional, non-convex, and stochastic black-box objectives arising from physics-based rollouts. At iteration $t$, CMA-ES samples candidate designs $d \sim \mathcal{N}(\mu_t, \Sigma_t)$ from a multivariate Gaussian distribution, evaluates their fitness score $J(d)$, and updates the distribution parameters $(\mu_t, \Sigma_t)$ via rank-based maximum-likelihood adaptation to bias sampling toward high-performing regions. This iterative distribution update process results in the best estimate of the optimal design $d^*$.

\begin{algorithm}[t]
\caption{Bi-Level End-Effector Co-Design}
\label{alg:codesign}
\KwIn{$E$ simulation environments, object SDF $\phi$, finger mesh $\rho_d$, training budgets $B_d,B_p$.}
\KwOut{Best design $d^\star$ and its score $J(d^\star)$.}

Initialize derivative-free design optimizer $\mathcal{O}_d$ over $\mathcal{D}$\;

\For{$t=1$ \KwTo $B_d$}{
  $d_t \leftarrow \mathcal{O}_d.\textsc{Suggest}()$\;
  Sample contexts $\{\xi_e\}_{e=1}^E$ and reset $E$ environments with $(d_t,\xi_e)$\;
  \ForPar{$e=1$ \KwTo $E$}{
    $\{p_{e,i}\}_{i=1}^I \leftarrow \textsc{CMA-ES}\!\left(f(\cdot;d_t,\xi_e),\mathcal{X},B_p\right)$\;
    \For{$i=1$ \KwTo $I$}{
      $J_{e,i}(d_t; p_{e,i}) \leftarrow \mathcal{S}.\textsc{Rollout}(d_t,\xi_e,p_{e,i})$\;
    }
    $p^\star_e \leftarrow \arg\max_{p_{e,i}}\ J_{e,i}(d_t; p_{e,i})$\;
    $J_e(d_t) \leftarrow \mathcal{S}.\textsc{Rollout}(d_t,\xi_e,p^\star_e)$\;
  }
  $J(d_t) \leftarrow \frac{1}{E}\sum_{e=1}^E J_e(d_t)$\;
  $\mathcal{O}_d.\textsc{Observe}(d_t,J(d_t))$\;
}
\Return{${d}^* = \mathcal{O}_d.\textsc{GetBest}()$}\;
\end{algorithm}

\subsection{Sim-to-Real Transfer via Policy Distillation}
After obtaining the optimized design $d^*$ and its privileged design-conditioned controller $\pi^*_{d,\text{prv}}$ in simulation, we distill this policy into a reactive controller using imitation learning. Specifically, we roll out $\pi^*_{d,\text{prv}}$ with the optimized design $d^*$ and collect $L=100$ expert trajectories per task, storing both control actions and synchronized object pointcloud observations (Fig.~\ref{fig:overview}-d). These demonstrations are used to train a student 3D diffusion policy $\pi^*_{d,\text{pcd}}$~\cite{ze20243d,chi2025diffusion} based on pointcloud (pcd) observations.

At each timestep, the student policy receives a non-privileged observation consisting of the end-effector configuration $p=(\alpha, R, w)$, the segmented object point cloud, and its centroid, and outputs an action $a_t = (\Delta \alpha, \Delta R, \Delta w)$ corresponding to changes in Cartesian position, orientation in axis-angle form, and gripper width commands. We adopt a 3D diffusion policy architecture due to its strong performance under limited demonstration data and robustness to multimodal action distributions~\cite{ze20243d}. During deployment (Fig.~\ref{fig:overview}-e), the distilled policy $\pi^*_{d,\text{pcd}}$ is executed zero-shot on the real robot using RGB-D pointcloud observations, preserving the gentle manipulation behavior learned in simulation.

\section{Evaluation}
\label{sec:evaluation}
We evaluate the proposed co-design framework to answer the following research questions (RQs): 
(1) Does the proposed diffeomorphic design space enable diverse and physically meaningful end-effector geometries? 
(2) Do the co-designed end-effectors perform well across tasks and object variations? 
(3) How does our approach compare against standard parallel-jaw grippers and baseline design methods? 
(4) Can the design and control strategies learned in simulation be transferred effectively to real-world execution?

To this end, we design three representative DFOM tasks: jelly grasping, jelly pushing, and fillet scooping, with the first two also evaluated in the real world. Jelly and raw fish fillet exhibit extremely low stiffness and yield stress, which we model in the physical simulator. These material properties make the tasks particularly challenging and suitable for evaluating gentle manipulation performance.

\subsection{Simulation Experiment}
We first evaluate the proposed method in simulation and compare against the following baselines:
(i) \textit{PJ}: a standard parallel-jaw gripper mounted on the xArm7 robot, which features flat contact pads;
(ii) \textit{BO}: replacing the design optimizer $\mathcal{O}_d$ with Bayesian Optimization, a commonly adopted approach in prior co-design work~\cite{bjelonic2023learning,kim2021mo}, while keeping the same design-conditioned control strategy as our method. We use the Lower Confidence Bound (LCB) acquisition function;
(iii) RL: end-to-end co-design via reinforcement learning (RL), formulated as a dual design–control Markov decision process problem~\cite{liu2023learning,ikemuraefficient}. We compare against the method of~\cite{ikemuraefficient}, using Reinforcement Learning from Prior Data (RLPD)~\cite{ball2023efficient} as the backbone. For this setup, we collect 20 human demonstration trajectories for each task, where each episode is executed with a randomly sampled design $d \in \mathcal{D}$.

All simulation experiments are conducted in Genesis using its built-in Taichi-based soft-body physics engine~\cite{hu2020difftaichidifferentiableprogrammingphysical}. We adopt the Material Point Method (MPM) instead of alternatives such as finite element methods, as MPM provides improved numerical stability and accuracy under the large deformations typical in DFOM scenarios~\cite{de2020material}. In this formulation, deformable objects are represented as particle-based soft bodies, as visualized in Fig.~\ref{fig:design-evolution}. 
All quantitative simulation results (for Ours, PJ, BO and RL) are averaged over three random seeds, with mean and standard deviation reported.

\subsubsection{Co-Design performance across tasks}
We evaluate the performance of the optimized designs $d^*$ across tasks and object geometries in simulation to address RQ2. For grasping and pushing, we consider two representative object shapes: a cylindrical jelly object with a curved contact surface, and an ``onigiri''-shaped triangular prism whose non-parallel side faces pose challenges for standard parallel-jaw grippers. For grasping, we additionally evaluate a large jelly cube to assess applicability to bulky objects. For scooping, we use a flat rectangular fillet, which requires stable surface contact and precise force distribution.

To improve robustness, we apply domain randomization over object material properties and task configurations. Specifically, Young's modulus and Poisson's ratio are randomized within ranges that cover the estimated physical characteristics of the real-world materials. In addition, we randomize the object's initial pose and size, the robot's initial end-effector configuration, and inject Cartesian action noise during execution.
Performance is measured using three metrics: the overall design score $J(d^*)$, task success rate $q_{\text{succ}}$, and maximum stress $\sigma^{\mathrm{max}}$ (with $s=2.5$ in Eqs.~\ref{eq:max-stress-spatial} and~\ref{eq:stress-temporal}). All metrics are computed over 50 rollout trials and reported as averages.

\begin{table*}[!htp]\centering
\caption{Optimal design performance of the jelly grasping task}
\label{tab: quant-grasp}
\begin{tabular}{lccc|ccc|ccc}\toprule
& \multicolumn{3}{c}{cylinder} 
& \multicolumn{3}{c}{onigiri} 
& \multicolumn{3}{c}{large cube} \\\cmidrule(lr){2-4}\cmidrule(lr){5-7}\cmidrule(lr){8-10}

& score $\uparrow$ & success $\uparrow$ & stress $\downarrow$
& score $\uparrow$ & success $\uparrow$ & stress $\downarrow$
& score $\uparrow$ & success $\uparrow$ & stress $\downarrow$ \\\midrule

Ours 
& \textbf{143} $\pm$ \textbf{7} 
& \textbf{0.97} $\pm$ {0.03} 
& \textbf{9274} $\pm$ 124 
& \textbf{64} $\pm$ 18 
& \textbf{0.73} $\pm$ 0.08 
& 13973 $\pm$ 490 
& 19 $\pm$ 48 
& \textbf{0.51} $\pm$ 0.21 
& \textbf{12050} $\pm$ 1116 \\

PJ 
& 79 $\pm$ 12 
& 0.63 $\pm$ 0.05 
& 9366 $\pm$ \textbf{120} 
& -33 $\pm$ \textbf{3} 
& 0.12 $\pm$ \textbf{0.04} 
& \textbf{10224} $\pm$ \textbf{332} 
& -133 $\pm$ \textbf{17} 
& 0.18 $\pm$ \textbf{0.04} 
& 18893 $\pm$ \textbf{620} \\

BO 
& 106 $\pm$ 28 
& 0.80 $\pm$ 0.11 
& 9998 $\pm$ 1089 
& 30 $\pm$ 66 
& 0.59 $\pm$ 0.27 
& 14946 $\pm$ 1658 
& \textbf{33} $\pm$ 53 
& 0.45 $\pm$ 0.40 
& 13189 $\pm$ 1030 \\

RL 
& -110 $\pm$ 62	 & 0.04 $\pm$ \textbf{0.02}	 &  22015 $\pm$ 12105	
& -244 $\pm$ 97	 & 0.09$\pm$0.06	 &  38107$\pm$	19327
&  - &  -& - \\

\bottomrule
\end{tabular}
\end{table*}

\paragraph{Jelly Grasping}
Quantitative results are summarized in Table~\ref{tab: quant-grasp}. Across object geometries and evaluation metrics, our method achieves the best overall performance compared to the baselines. BO provides a competitive alternative, but is less effective than evolutionary strategies in exploring the high-dimensional latent design space ($M_{\text{latent}} > 10$), resulting in less effective end-effector geometries in several scenarios.

\begin{figure}[t]
\centering
\includegraphics[width=0.95\linewidth]{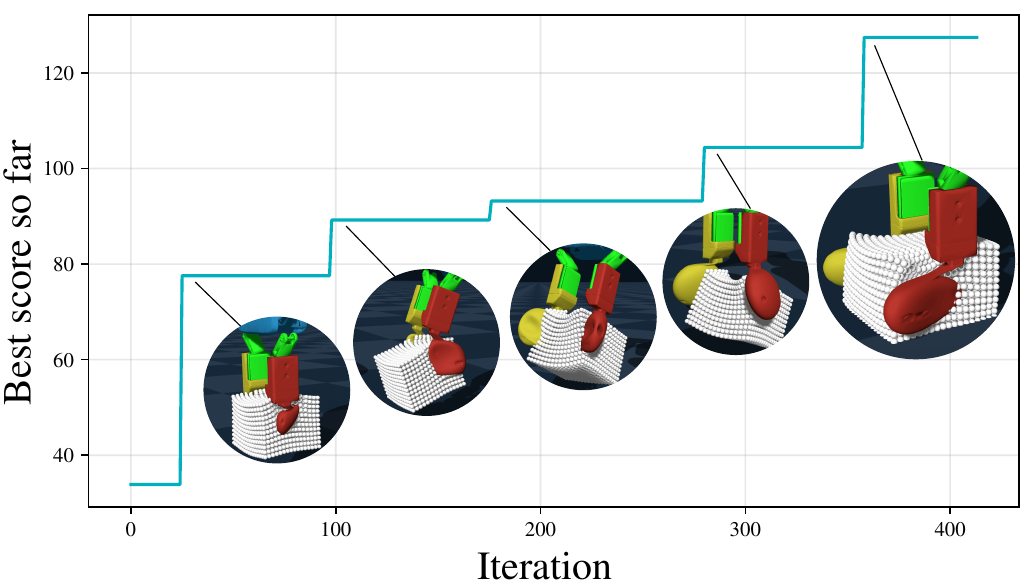}
\caption{End-effector design evolution for grasping a large jelly cube during a typical optimization run. High-scoring designs exhibit larger contact surfaces, which improve grasp stability and reduce contact-induced stress.}
\label{fig:design-evolution}
\end{figure}

Beyond relatively simple shapes such as cylinders and triangular prisms, we further evaluate the grasping of a large jelly cube with side length exceeding $8.5$\,cm, which is the maximum width of the original gripper. The corresponding design evolution is shown in Fig.~\ref{fig:design-evolution}. The optimization process converges toward geometries with increased contact surface area and outward-extending structures, which improve grasp stability and distribute contact forces more evenly to accommodate bulky fragile objects.
These results directly address RQ3 and highlight the necessity of co-design for large-object DFOM tasks, where standard parallel-jaw grippers are constrained by limited opening width and contact area. As shown in Table~\ref{tab: quant-grasp}, the co-designed end-effector achieves a 33\% higher success rate and reduces induced stress by 36\% compared to the parallel-jaw baseline, which struggles to reliably grasp the oversized fragile objects.

Notably, RL performs poorly, exhibiting both a low success rate and excessively high stress metrics. We attribute this to two primary factors. First, identifying an appropriate solution in the joint design–control space within a single training run is extremely challenging, particularly under a limited sample budget, as deformable body simulation is computationally expensive and time-consuming. Second, although we incorporate stress-related penalties to encourage gentle manipulation, their weight must be tuned down in practice, as stronger penalties induce reward hacking, causing the agent to avoid interacting with the object to minimize negative rewards. As illustrated in the Appendix, the RL agent typically adopts an overly tight grasp. In contrast, our framework alleviates both issues by employing a motion-adaptive, design-conditioned policy and leveraging it for the outer-loop design search.

\begin{table*}[!htp]\centering
\caption{Optimal design performance of the jelly pushing task}
\label{tab: quant-push}
\begin{tabular}{lccc|ccc}\toprule
& \multicolumn{3}{c}{cylinder} 
& \multicolumn{3}{c}{onigiri} \\\cmidrule(lr){2-4}\cmidrule(lr){5-7}
& score $J$ & success rate $q_{\text{succ}}$ & stress $\sigma^{\mathrm{max}}$
& score $J$ & success rate $q_{\text{succ}}$ & stress $\sigma^{\mathrm{max}}$ \\\midrule
Ours 
& \textbf{85} $\pm$ \textbf{5} 
& \textbf{0.760} $\pm$ \textbf{0.000} 
& \textbf{1236} $\pm$ 108 
& \textbf{70} $\pm$ \textbf{16} 
& \textbf{0.627} $\pm$ \textbf{0.076} 
& \textbf{1177} $\pm$ 85 \\
BO 
& 63 $\pm$ 16
& 0.633 $\pm$ 0.064 
& 1289 $\pm$ \textbf{46} 
& 65 $\pm$ 23
& \textbf{0.627} $\pm$ 0.115 
& 1295 $\pm$ \textbf{64} \\
RL 
&  -39	$\pm$ 31	&  0.280$\pm$ 	0.100	&  2432$\pm$ 	522	&  -81$\pm$ 	33	&  0.287$\pm$ 	0.163&  	5032$\pm$ 	659 \\

\bottomrule
\end{tabular}
\end{table*}

\paragraph{Jelly pushing}
Results in Table~\ref{tab: quant-push} show that our method achieves a 6.4\% higher success rate and a 6.6\% reduction in induced stress on average across test objects compared to BO. Interestingly, the optimized end-effector geometry converges to a shape resembling the head of a golf club (Fig.~\ref{fig:real-ee-vis}), featuring a concave contact surface that partially constrains the object during pushing. This geometric structure improves contact stability and enhances robustness to pose and contact uncertainties during planar pushing. As in the grasping task, the RL baseline remains suboptimal; however, because pushing is less challenging, its performance degradation is less severe.

\begin{table}[!tp]\centering
\caption{Optimal design performance of the fillet scooping task}
\label{tab: quant-scoop}
\begin{tabular}{lccc}\toprule
 & score $J \uparrow$  & success rate~$q_{\text{succ}} \uparrow$ & stress~$\sigma^{\mathrm{max}} \downarrow$  \\\midrule
Ours & \textbf{293 $\pm$ 97} & \textbf{0.779 $\pm$ 0.203} & 17154 $\pm$ \textbf{469} \\
BO & 282 $\pm$ 146 & 0.683 $\pm$ 0.322 & \textbf{13586} $\pm$ 2857 \\
\bottomrule
\end{tabular}
\end{table}
 
\begin{figure}[t]
\centering
\includegraphics[width=0.9\linewidth]{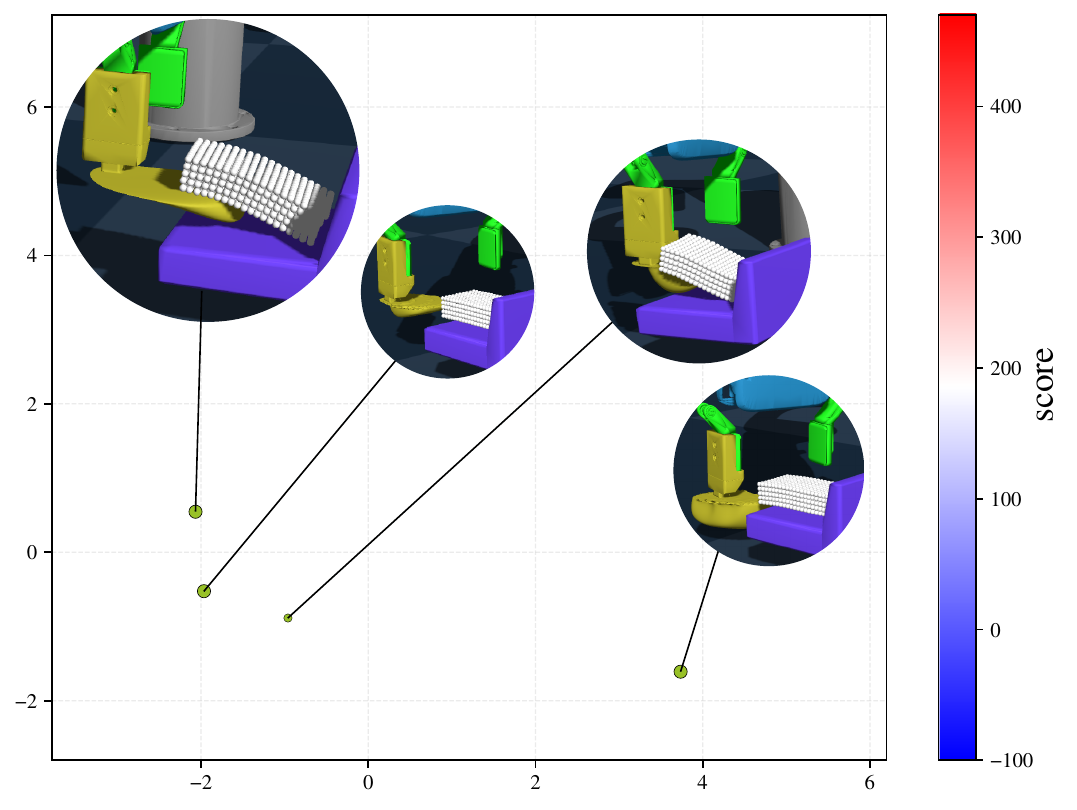}
\caption{Visualization of the design-space landscape projected onto two principal components from the $M_{\text{latent}}$-dimensional latent diffeomorphic space. Warmer colors indicate higher end-effector performance for the fillet scooping task. It suggests that thinner and longer end-effector geometries tend to perform better for scooping.}
\label{fig:design-heatmap}
\end{figure}

\paragraph{Fillet Scooping}
Since the LDM design space spans a wide range of geometries, effective scooping designs only exist in a task-specific subregion. As Fig.~\ref{fig:design-heatmap} illustrates, the high-performing region is concentrated in the lower-left area of the latent landscape, corresponding to longer and thinner geometries that can slide underneath the fillet while avoiding collisions with the supporting surface. Table~\ref{tab: quant-scoop} shows that our method achieves higher overall scores and task success rates compared to baselines, but also induces higher stress in some cases. This occurs primarily when the scoop is inserted deeply and compresses the fillet against the table, indicating that further improvements in fine-grained contact control could reduce stress while preserving task performance. We omit RL in this task due to the complexity of reward engineering.

\begin{table}[!tp]\centering
\caption{Comparison of design spaces}
\label{tab: quant-design-spaces}
\begin{tabular}{lccc}\toprule
& score $J$ & success rate $q_{\text{succ}}$ & stress $\sigma^{\mathrm{max}}$ \\\midrule
Ours   & \textbf{64} $\pm$ 18  & \textbf{0.727 $\pm$ 0.081} & \textbf{13973} $\pm$ 490 \\
Cubic  & 10 $\pm$ \textbf{17}   & 0.473 $\pm$ 0.083          & 14894 $\pm$ \textbf{306} \\
Spherical & 13 $\pm$ 23           & 0.540 $\pm$ 0.106          & 15321 $\pm$ 386          \\
\bottomrule
\end{tabular}
\end{table}

\subsubsection{Design space comparison} 
We include simple, primitive shape spaces as low-dimensional parametric baselines to evaluate the effect of design-space expressiveness on co-design performance. Two shape primitive design spaces are used for this purpose, cubic shapes parameterized by side length and spherical shapes parameterized by radius. Results on the onigiri grasping task are summarized in Table~\ref{tab: quant-design-spaces}. The LDM representation consistently outperforms primitive design spaces in both task success and stress reduction. In contrast, primitive geometries lack sufficient expressiveness to adapt contact geometry to object shape, leading to unstable grasps and excessive localized stress, particularly when interacting with non-parallel object surfaces. These results directly address RQ1 and demonstrate that LDM enables physically meaningful and task-effective end-effector design exploration.


\begin{table}[!tp]\centering
\caption{Real-World Experiments}
\label{tab: quant-real-world}
\begin{tabular}{llcc}\toprule
Task & Method & success rate $\uparrow$ & object damage $\downarrow$ \\\midrule

\multirow{2}{*}{Grasp}
& Ours & \textbf{9}/9 & \textbf{1}/9 \\
& BO   & {5/9} & \textbf{1}/9 \\

\midrule

\multirow{2}{*}{Push}
& Ours & \textbf{10}/12 & \textbf{0}/12 \\
& BO   & {9/12} & \textbf{0}/12 \\

\bottomrule
\end{tabular}
\end{table}

\subsection{Real-World Experiment}
Real-world experiments are conducted on a UFactory xArm 7, a 7-DoF robotic manipulator. An external Intel RealSense L515 LiDAR camera is mounted in front of the robot to capture object point clouds. All co-designed end-effectors are 3D printed and attached to the robot’s standard parallel-jaw gripper. The jelly objects are fabricated using gelatin cast in 3D-printed molds and dyed with green pigment to facilitate point-cloud segmentation. Evaluation metrics include binary task success and a binary object integrity indicator based on visible breakage of the jelly. We evaluate both grasping and pushing of cylindrical jelly objects using our method and compare against the BO baseline.

\begin{figure}
    \centering
    \includegraphics[width=0.7\linewidth]{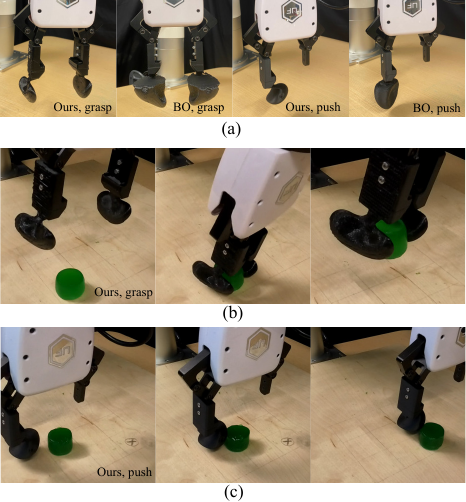}
    \caption{(a) Task-specific end-effector designs optimized, fabricated, and mounted on the robot for real-world evaluation. Two design optimizers $\mathcal{O}_d$ are compared: CMA-ES (ours) and BO for jelly cylinder grasping and pushing tasks. The co-designed end-effector exhibits curved inner contact surfaces that improve form closure and reduce contact-induced damage to fragile objects. (b) \& (c) The roll out of our policy and BO baseline in real world.}
    \label{fig:real-ee-vis}
\end{figure}

Four sets of task-specific end-effector designs obtained from the two optimization methods are tested (Fig.~\ref{fig:real-ee-vis}). For grasping, the optimized design exhibits a curved and concave inner contact surface that increases contact area and promotes force and form closure, improving grasp stability while reducing stress concentration. For pushing, the optimized geometry resembles a golf-club-shaped head with a curved contact profile, which improves directional control and robustness to contact uncertainty. Results in Table~\ref{tab: quant-real-world} address RQ4 and demonstrate that the distilled diffusion policy, operating on pointcloud observations and robot proprioception, transfers successfully to the real robot. Additional experimental details are provided in the appendix.


\section{Discussion and Limitation}

This work demonstrates that computational co-design can effectively shift part of the task complexity from control to hardware. Although the proposed design-conditioned motion-adaptive controller is not globally optimal, jointly optimizing morphology enables the discovery of end-effector geometries that compensate for controller limitations. This observation aligns with the principle of \emph{morphological computation}~\cite{paul2006morphological}, where physical structures implicitly perform functions that would otherwise require complex feedback control. In practice, this suggests that intelligent morphology can substantially reduce control complexity in challenging DFOM tasks.



Despite its effectiveness, the proposed framework has several limitations. First, design data collection and motion primitive design reduce sample complexity but still require manual tuning of controller hyperparameters and task-specific configurations. Second, sim-to-real discrepancies in object physical properties remain challenging. Factors such as material heterogeneity in food materials and temperature-dependent mechanical behavior can degrade real-world performance despite domain randomization. Addressing these gaps through improved material modeling and online adaptation remains an important direction for future work.


\section{Conclusion}
\label{sec:conclusion}
We present a data-driven co-design framework for deformable and fragile object manipulation that jointly optimizes end-effector morphology and manipulation strategy. Our results highlight two key findings. First, the proposed latent diffeomorphic design space enables expressive yet tractable optimization, producing diverse and task-effective end-effector geometries. Second, exploiting privileged physical signals from simulation, such as soft-body stress, allows the discovery of hardware designs that improve task success while substantially reducing contact-induced object damage. The proposed sim-to-real pipeline achieves robust performance across grasping and pushing tasks in both simulation and real-world experiments. Looking forward, extending the framework to multi-task robot co-design represents a promising direction for fresh food handling applications, such as designing end-effectors capable of both scooping and cutting fillets without requiring tool changes.




\begin{figure*}[t]
\centering
\includegraphics[width=0.95\linewidth]{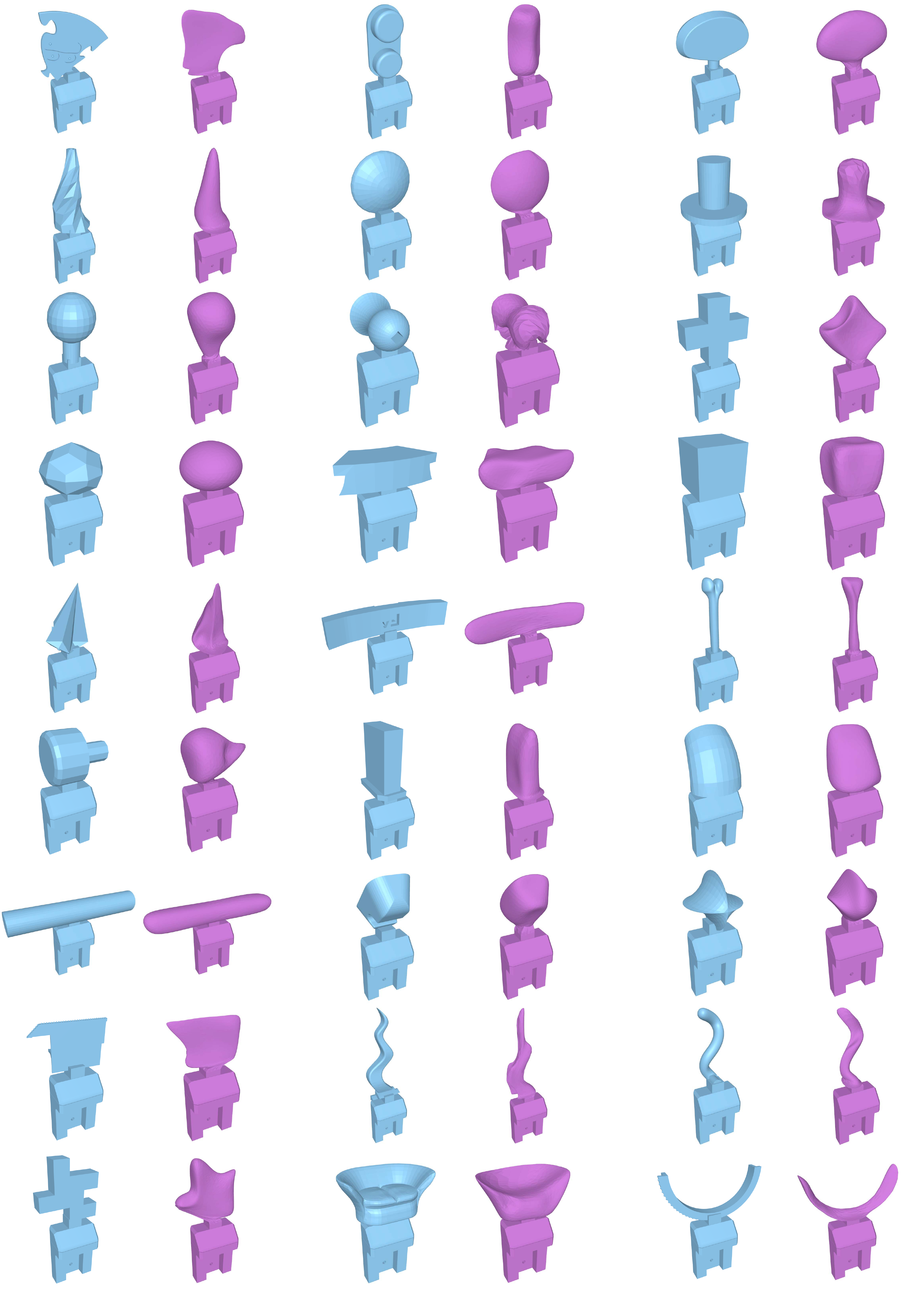}
\caption{Additional examples of diffeomorphic parameter fitting for jelly grasping and pushing. The target mesh (left, blue) and the fitted mesh (right, purple) closely match in terms of Chamfer distance. The target shapes are drawn from a subset of the $N_0$ designs used to learn the raw diffeomorphic parameters.}
\label{fig:design-target-fitted}
\end{figure*}

\begin{figure*}[t]
\centering
\includegraphics[width=0.95\linewidth]{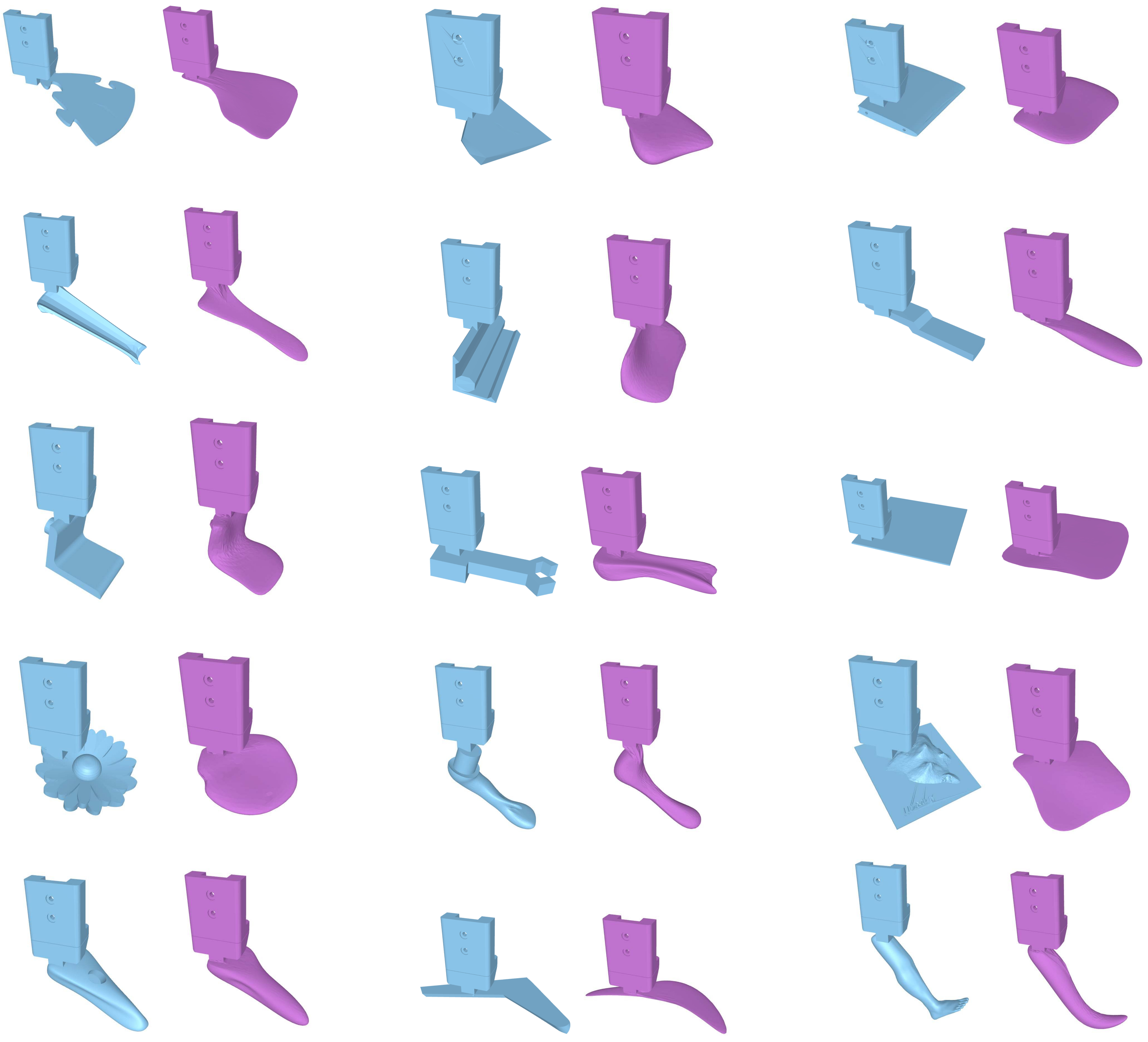}
\caption{Subset of the diffeomorphic training dataset and corresponding parameter fitting results for the fillet scooping task.}
\label{fig:design-target-fitted-scoop}
\end{figure*}

\begin{table*}[b!]
\centering
\caption{Comparison of co-design methods by design and action space dimensionality and paradigm.}
\label{tab:codesign_comparison}
\begin{tabular}{lccc}
\hline
\textbf{Paper} & \textbf{Design Space Dimension} & \textbf{Action Space Dimension} & \textbf{Co-Design Paradigm} \\
\hline
SERL Co-Design~\cite{cheng2024structural} & 2 & 9 & Two-stage: RL + GA \\
Meta-RL Co-Design~\cite{belmonte2022meta} & 4 & 16 & Two-stage: RL + CMA-ES \\
Object Adaptation~\cite{guo2024learning} & 4 & 7 & Single-stage: Dual MDP \\
Object Adaptation~\cite{guo2024learning} & 4 & 7 & Single-stage: Dual MDP \\
Cross-Embodiment Hand Co-Design~\cite{fay2025cross} & 5 & 20 & Two-stage with Design-Conditioned Policy \\
CageCoOpt~\cite{dong2025cagecoopt} & 6 & 4 & Two-stage: PPO + BO/GA \\
Evolving Robot Hand~\cite{yang2024evolving} & 8 & 26 & Two-stage: RL + Gradient-based Optimization \\
MORPH~\cite{he2024morph} & 10 & 11 & Single-stage RL + Neural Surrogate \\
Tool Design and Use~\cite{liu2023learning} & 12 & 6 & Single-stage: PPO \\
HWasP~\cite{chen2020hardware} & 12 & 9 & Single-stage: TRPO \\
EvoGym~\cite{bhatia2021evolution} & 15 & 35 & Two-stage: PPO + BO/GA \\
Co-Design with CBD~\cite{xu2021end} & 17 & 8 & End-to-end Differentiable Co-Design \\
Soft Gripper Co-Design~\cite{yi2025co} & 22 & 8 & Neural-physics surrogate \\
NGE~\cite{wang2019neural} & $>2$ & 10 & Two-stage with Evolving Controller \\
\hline
\end{tabular}
\end{table*}

\clearpage
\appendix

\subsection{Additional experiment results and details}
\label{app:experiment}
\begin{figure*}[t]
\centering
\includegraphics[width=0.99\linewidth]{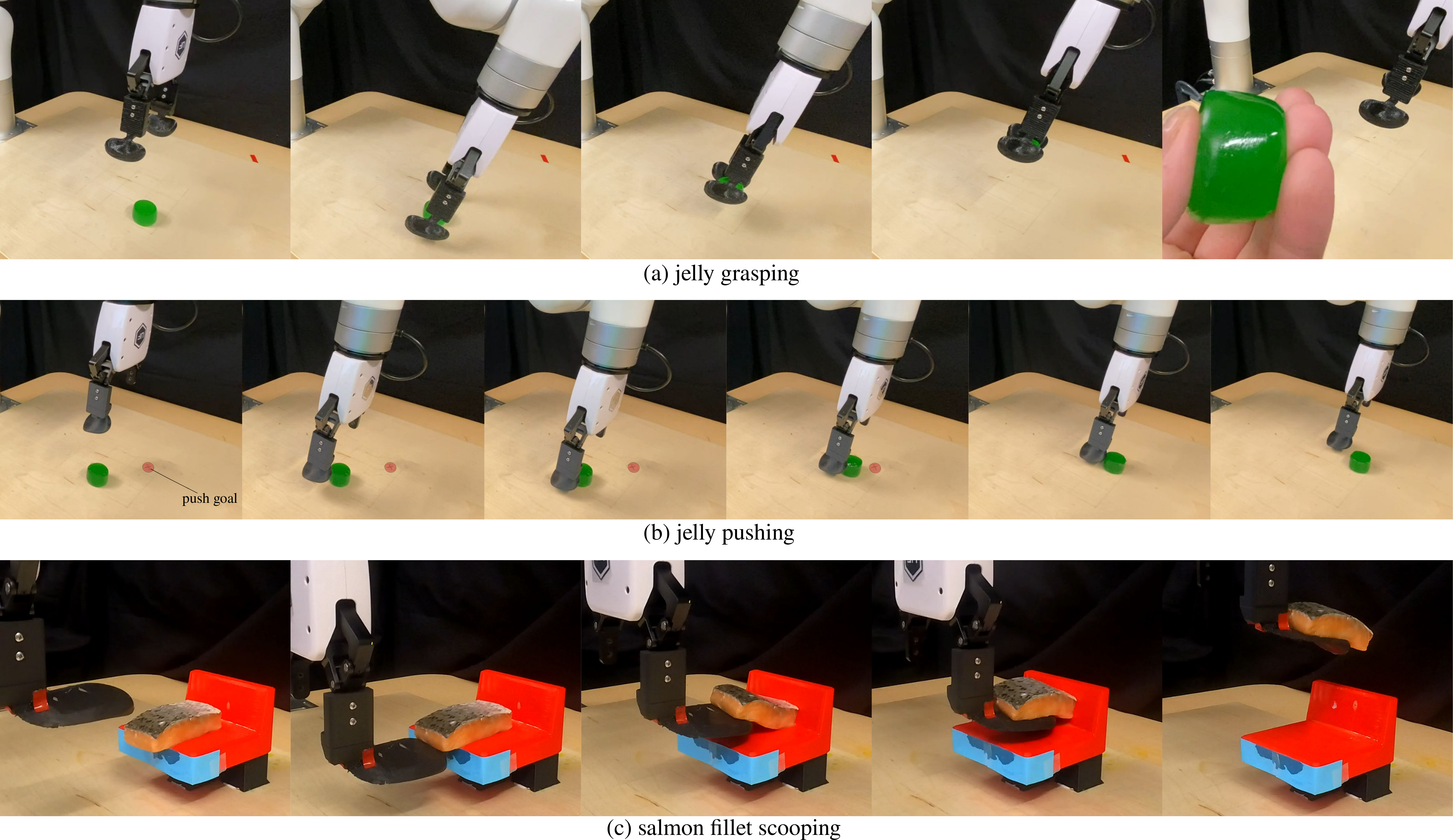}
\caption{Real-world rollouts of the co-designed policies on the three DFOM tasks.}
\label{fig:real-world-trajs}
\end{figure*}

We show real-world rollouts of the co-designed end-effectors for the jelly grasping and pushing tasks in Fig.~\ref{fig:real-world-trajs}. We additionally include the salmon fillet scooping task, which was omitted from the main paper due to page limits. Further qualitative results, including both successful and failed episodes, are provided in the supplementary video.

\begin{figure}[b]
\centering
\includegraphics[width=0.99\linewidth]{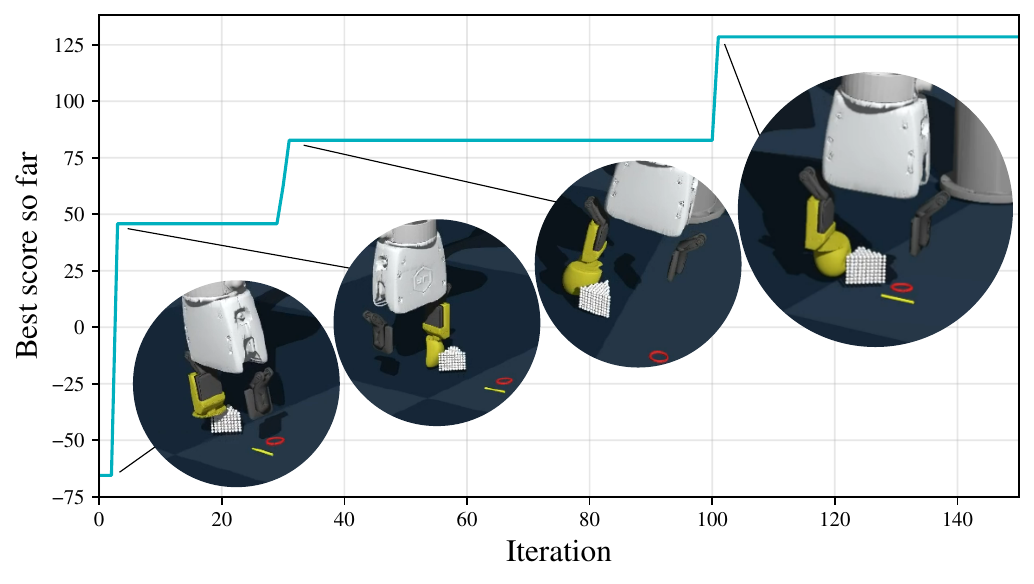}
\caption{End-effector design evolution for pushing a jelly onigiri during a typical optimization run. High-scoring designs exhibit larger contact surfaces, which improve push stability.}
\label{fig:design-evolution-push}
\end{figure}

The design evolution for the jelly onigiri pushing task is shown in Fig.~\ref{fig:design-evolution-push}. During optimization, the end-effector geometry evolves toward increased contact surface area. This is particularly beneficial because the onigiri can pivot and roll about its edges upon contact; a larger contact surface mitigates rolling and helps maintain stable control during pushing.

\begin{figure}[b]
\centering
\includegraphics[width=0.95\linewidth]{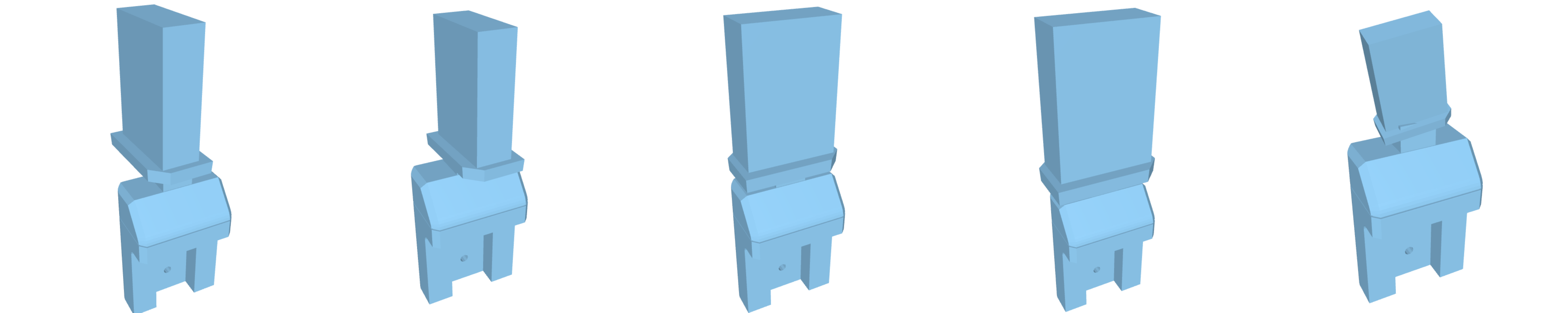}
\caption{Example base shape in the LDM training dataset (leftmost) and its augmented variants generated via random rotation, translation, and scaling.}
\label{fig:ldm-augmented}
\end{figure}

\begin{figure}[t]
\centering
\includegraphics[width=0.9\linewidth]{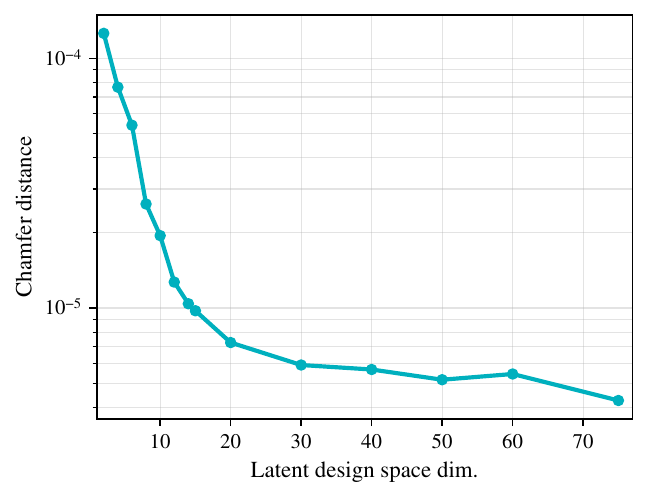}
\caption{Average Chamfer distance between original mesh and fitted mesh, with respect to latent design space dimension~$M_{\text{latent}}$. We choose $M_{\text{latent}}=15$ in most of the experiments as a tradeoff.}
\label{fig:chamfer_dist}
\end{figure}

For the fish fillet manipulation tasks, real objects are preprocessed into approximately cubic shapes with small geometric variations. 

For the jelly grasping task, we employ symmetric, co-designed end-effector fingers. Symmetry reduces the effective design complexity. It is sufficient for the relatively regular object shapes considered in this work. Exploring asymmetric finger designs is a promising direction for future work, as it may enable handling objects with greater geometric diversity and more complex local shape features.

\subsection{Design Space Parametrization}
\label{app:design-space}
We apply early stopping when fitting diffeomorphic design parameters in the gradient descent: if no improvement is observed for 320 iterations after the first $10\%$ of the total optimization budget, the fitting process is terminated.

\begin{figure*}[t!]
\centering
\includegraphics[width=0.99\linewidth]{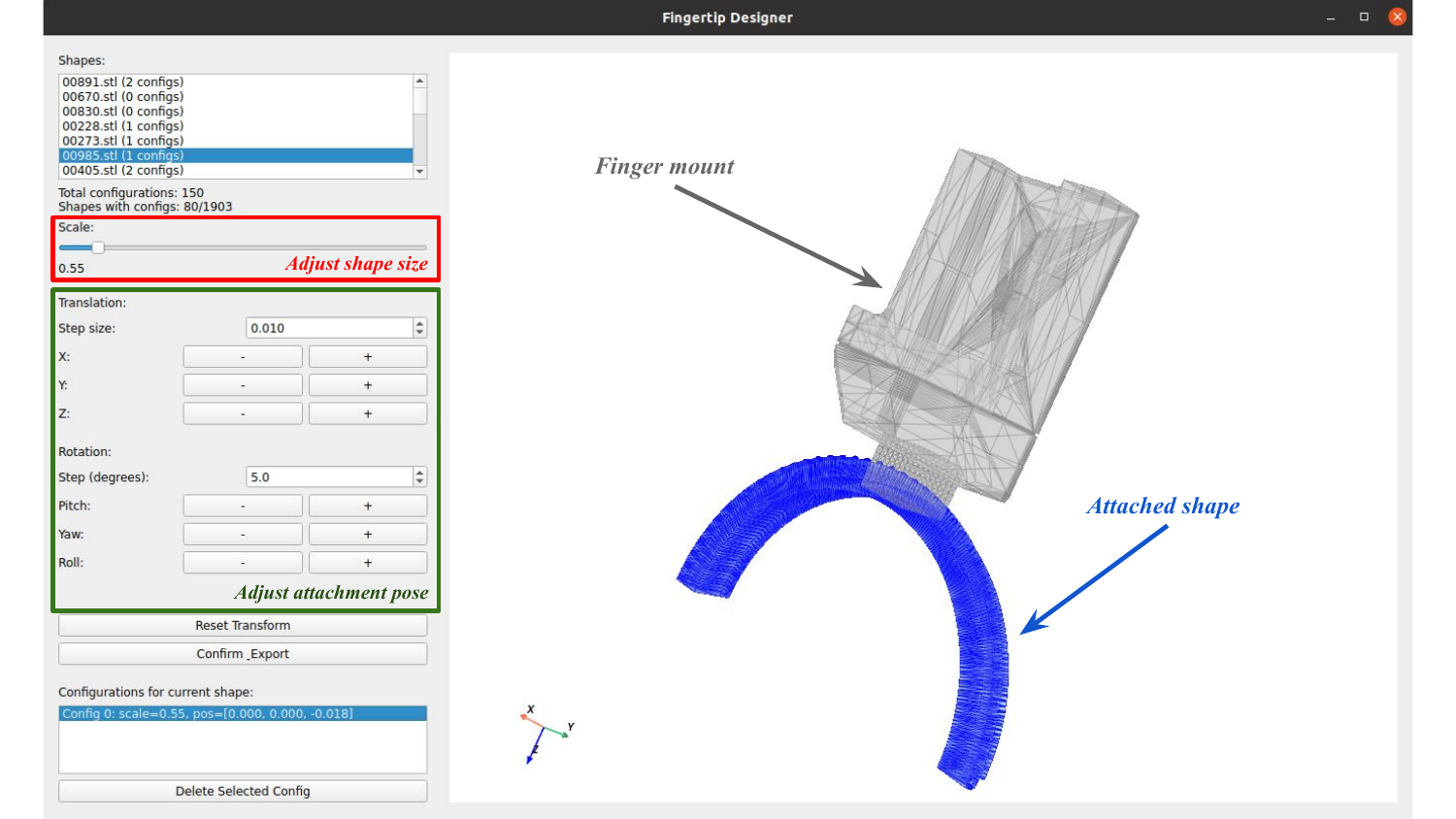}
\caption{Graphical user interface for curating end-effector designs used in diffeomorphic design space learning. Shapes from Thingi10K are attached to a standard finger mount at a user-specified pose.}
\label{fig:gui}
\end{figure*}

We visualize representative target meshes from the dataset used to learn the latent diffeomorphic parameters, along with their corresponding fitted meshes, in Fig.~\ref{fig:design-target-fitted} (for grasping and pushing) and Fig.~\ref{fig:design-target-fitted-scoop} (for scooping). These shapes are manually curated by attaching meshes from Thingi10K to a fixed connector compatible with the xArm7 standard end-effector at a specified pose using the GUI shown in Fig.~\ref{fig:gui}, yielding an initial dataset of $N_0 = 150$ designs. We then augment this set to a total of $N = 1000$ shapes by randomly rotating, translating, and scaling the original $N_0$ designs, as illustrated in Fig.~\ref{fig:ldm-augmented}. We use one such dataset for the jelly grasping and pushing tasks, and a separate dataset of the same size $N$ for the scooping task, which requires a higher proportion of flat geometries.

We choose the latent design space dimension $M_\text{latent} = 15$ based on the fitting accuracy shown in Fig.~\ref{fig:chamfer_dist}. When $M_\text{latent} < 15$, the representation lacks sufficient expressiveness to capture fine-grained geometric details. Increasing $M_\text{latent}$ beyond 15 yields diminishing improvements in fitting quality. We therefore select a lower-dimensional latent space to enable efficient sampling under a limited optimization time budget.

\subsection{Details in Bi-Level Design-Control Optimization}

\subsubsection{Design-Conditioned Control Optimization Objective}
\label{app:control-opt}
Given a fixed end-effector design $d$ and task context $\xi$, we synthesize a pre-contact configuration~$p^\star$ by maximizing a geometry-aware score that balances collision avoidance, proximity to the target object, and task-specific geometric alignment criteria:
\begin{equation}
p^\star
=
\arg\max_{p}
\;\;
f_{\text{pen}}(p)
+
f_{\text{near}}(p)
+
\sum_{m \in \mathcal{M}_{\text{task}}} f_m(p).
\end{equation}
Here $p$ denotes the candidate end-effector pose (position, orientation, and gripper opening), evaluated using signed distance field (SDF) queries between the end-effector mesh and the environment geometry.

\paragraph{Penetration avoidance score}
$f_{\text{pen}}$ assigns high scores to collision-free configurations and penalizes interpenetration with the object, table plane, and auxiliary fixtures (e.g., boards in scooping). It is computed from SDF samples on the end-effector surface as
\begin{equation}
f_{\text{pen}}(p)
=
-
\sum_i \max(0, -\phi_i(p) + \delta),
\end{equation}
where $\phi_i$ denotes the SDF value at surface sample $i$ and $\delta$ is a small slack margin. This formulation rewards configurations that remain outside the object while softly tolerating near-contact states.

\paragraph{Nearness score}
$f_{\text{near}}$ encourages the end-effector to approach the target object and relevant support surfaces without inducing penetration. It is defined using truncated positive SDF values:
\begin{equation}
f_{\text{near}}(p)
=
-
\sum_i \max(\phi_i(p), 0),
\end{equation}
which assigns higher scores to configurations that are close to contact while remaining collision-free.

\paragraph{Task-specific objectives}
The set $\mathcal{M}_{\text{task}}$ contains additional geometric alignment scores tailored to each manipulation primitive:

\begin{itemize}
\item \textit{Grasping.}
We encourage antipodal contact geometry by rewarding opposing surface normals at the two finger contact regions, estimated via SDF gradients. Additional alignment terms encourage consistency between finger normals and the closure direction, favoring force-closure and stable grasp initiation.

\item \textit{Pushing.}
We reward alignment between the finger contact plane normal and the object-to-subgoal direction, and encourage the finger centroid to lie at a desired pre-push offset behind the object along the pushing axis.

\item \textit{Scooping.}
We enforce geometric alignment between the finger principal axis and the scooping direction, encourage lateral centering relative to the object, and reward vertical and longitudinal placement that enables sliding underneath the object while avoiding collision with the support surface.
\end{itemize}

\begin{figure*}[t]
\centering
\includegraphics[width=0.95\linewidth]{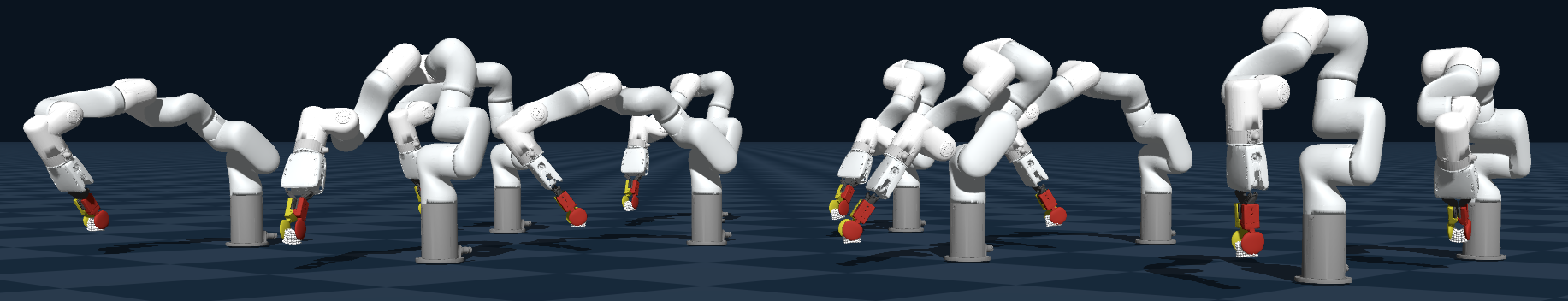}
\caption{Parallel, independent simulation environments in Genesis.}
\label{fig:parallel-envs}
\end{figure*}

Together, these terms define a pre-contact configuration scoring function that produces physically feasible and geometrically aligned initial configurations for closed-loop execution. Each design is evaluated in $E$ parallel environments with randomized task contexts $\xi_e$, where an independent configuration $p_e^\star$ is synthesized per environment (Fig.~\ref{fig:parallel-envs}), improving robustness to stochastic initialization effects.

\begin{figure}[t!]
\centering
\includegraphics[width=0.99\linewidth]{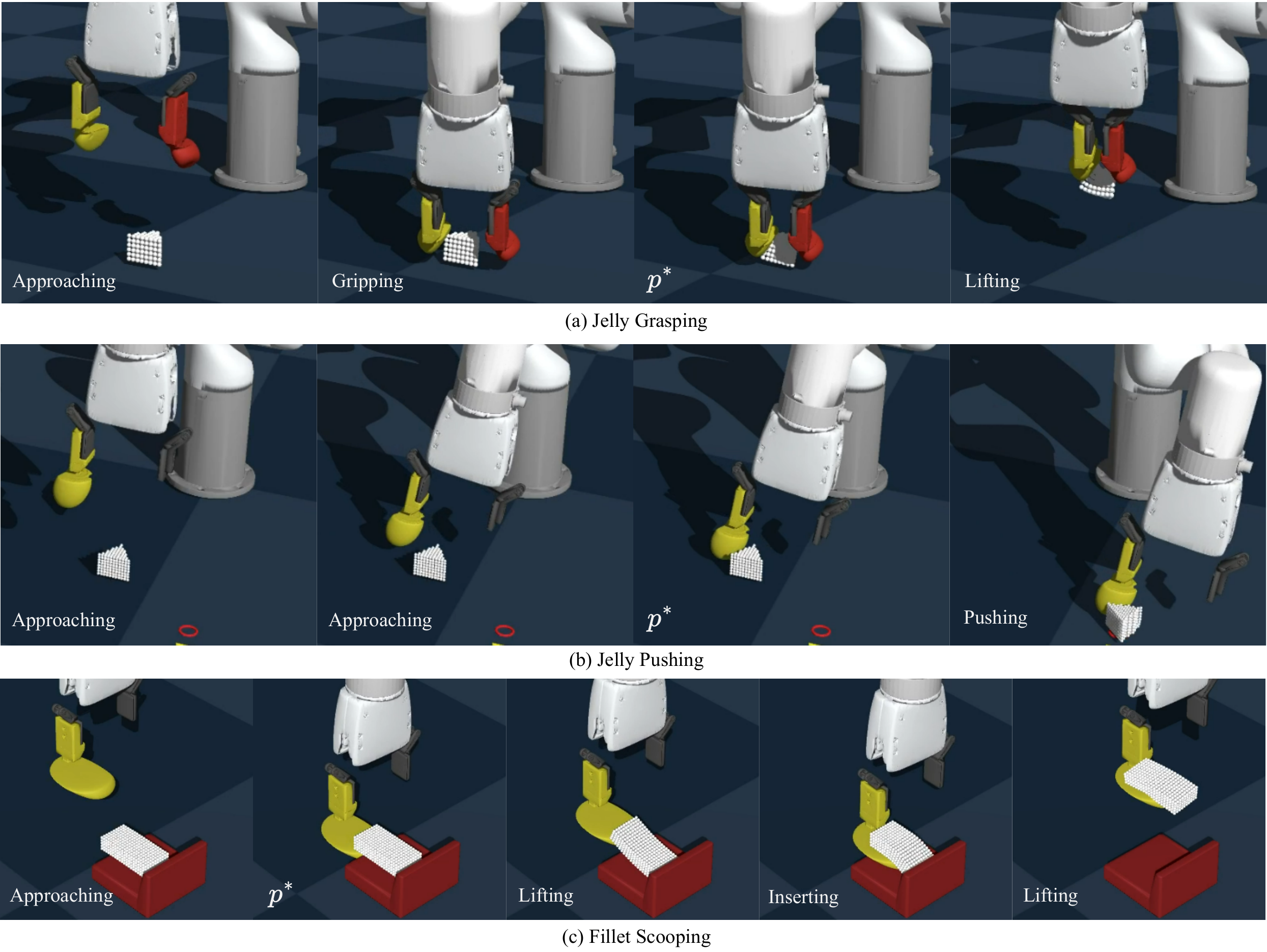}
\caption{Illustration of synthesized pre-contact configurations $p_d^\star$ and the corresponding motion primitive sequences for jelly grasping, pushing, and fillet scooping. Each row shows the ordered execution stages, including approach, interaction, and post-contact lifting, pushing or insertion.}
\label{fig:traj-tasks}
\end{figure}

\subsubsection{Motion Primitives}
This part is described in Section~\ref{paragraph:motion-primitive}. After obtaining the pre-contact configuration $p_d^\star$ for a given design sample $d$, which serves as a key geometric viapoint for task execution, our objective is to construct a design-conditioned execution policy $\pi_d^\star$ (short for $\pi_{d, \text{prv}}^\star$ as mentioned in Fig.~\ref{fig:overview}) that drives the system from the initial configuration $p_{\text{init}}$ to the task goal while passing through $p_d^\star$. Instead of learning a monolithic controller, we realize this objective using a sequence of motion primitives, as illustrated in Fig.~\ref{fig:traj-tasks}.

An execution rollout is represented as an ordered set of primitives
\begin{equation}
\pi_d^\star = \{\pi_{\text{approach}}, \pi_{\text{interaction}}, \pi_{\text{post-contact}}\},
\end{equation}
where each primitive $\pi_k$ produces control actions based on the current system state $s_t$ (encompassing privileged simulation signals such as robot's configuration $p$, object's centroid position and stress), the synthesized pre-contact pose $p_d^\star$, and the task context $\xi$:
\begin{equation}
a_t = \pi_k(s_t, p_d^\star, \xi), \quad k \in \{\text{approach}, \text{interaction}, \dots\}.
\end{equation}

Across all tasks, execution begins with an \emph{approach} primitive that uses collision-free trajectory planning to move the end-effector to the pre-contact configuration~$p_d^\star$. Then task-specific interaction primitives are activated. For grasping, this includes finger gripping and lifting with stress-aware stopping conditions. For pushing, a closed-loop planar controller drives the object toward the target location while adapting motion magnitude based on object displacement and contact feedback. For scooping, execution follows a sequence of lift--insert--lift primitives that regulate insertion depth and lifting height.

The proposed primitives are \emph{motion-adaptive} in that control signals are continuously modulated using privileged simulator feedback, including object pose, deformation state, and contact stress measurements. This enables robust execution under variations in object geometry, material properties, and initial pose across randomized environments. Furthermore, the modular primitive structure enables efficient large-scale trajectory generation for data collection, since primitives are lightweight, stable, and reusable across different end-effector designs and task instances.

\begin{figure}[t]
\centering
\includegraphics[width=0.95\linewidth]{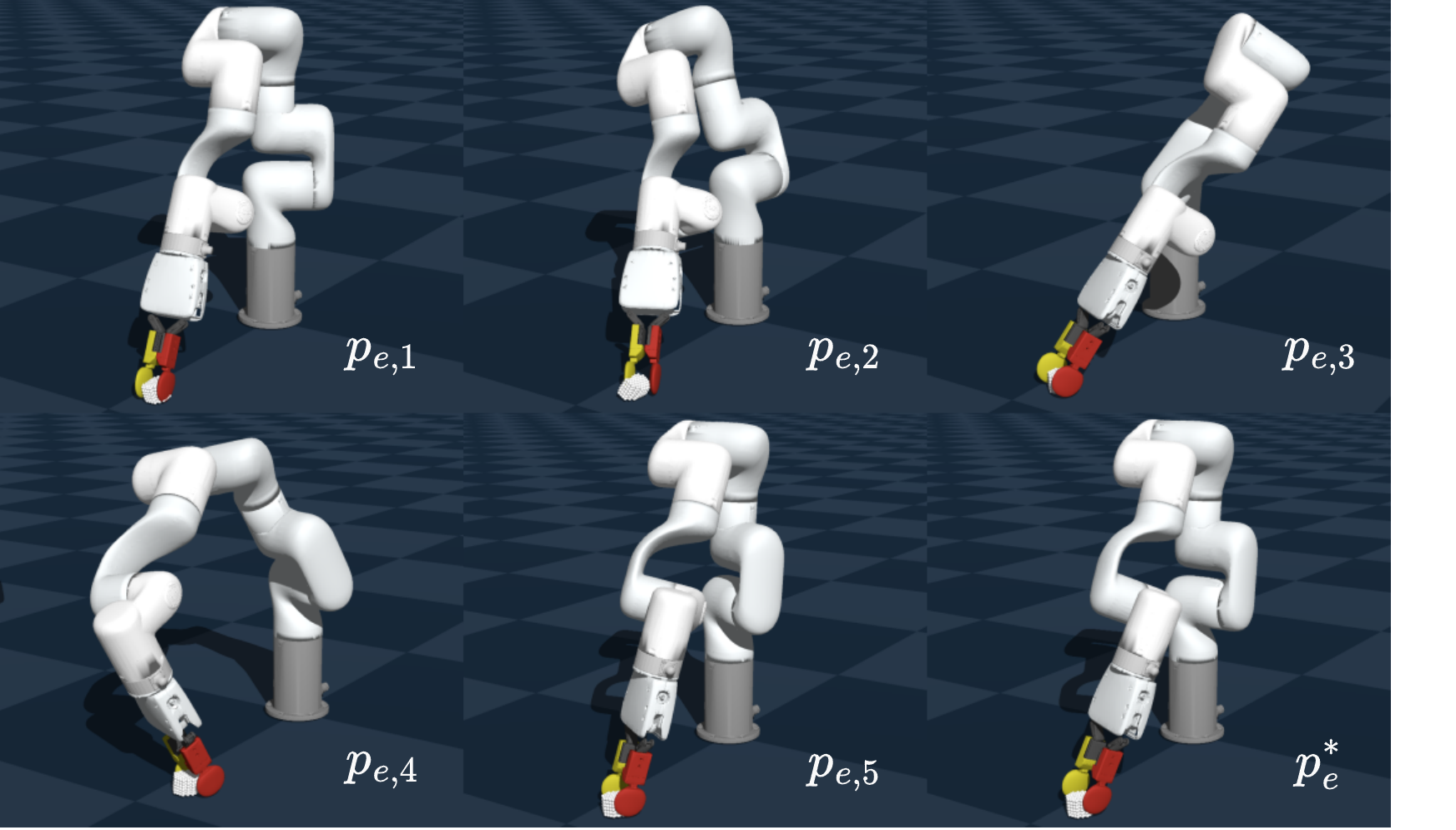}
\caption{illustration of the two-stage evaluation strategy in control optimization (line 6-9 in Algorithm~\ref{alg:codesign}). $p_{e,i}, i \in \{1,2,..,I\}$ refers to $I$ candidate pre-contact configurations synthesized from rigid body surrogate. Here $I=5$. $p^*_{e}$ refers to the best pre-contact configuration selected for execution.}
\label{fig:topk}
\end{figure}

In addition, we adopt a two-stage evaluation strategy, illustrated in Fig.~\ref{fig:topk}, to improve computational efficiency during control optimization. A small set of candidate pre-contact configurations is first generated using a rigid-body surrogate, after which the top-performing candidate is selected via soft-body simulation for execution.

\subsubsection{Design Optimization Objectives}
\label{app:design-opt}

For each candidate end-effector design $d$, we evaluate its performance by computing a score and aggregating across parallel simulation environments.

For a single environment rollout $e$, the evaluation score is defined as
\begin{equation}
J_e(d)
=
\lambda_{\text{prog}}\, q_{\text{prog}}
+ \lambda_{\text{succ}}\, q_{\text{succ}}
- \lambda_{\bar{\sigma}}\, \bar{\sigma}
- \lambda_{\mathrm{max}}\, \sigma^{\mathrm{max}},
\end{equation}

Below we define the progress and success metrics for each task. For simplicity, we omit the environment index $e$ and use discrete time indices $k \in \{1,\dots,K\}$.

\paragraph{Grasping and scooping}
Let $z_k$ denote the object centroid height at timestep $k$. The continuous progress metric is defined as the maximum achieved lift distance:
\begin{equation}
q_{\text{prog}} = \Delta z := \max_{k \in \{1,\dots,K\}} z_k - z_1.
\end{equation}

The binary success indicator requires the object to remain lifted above a threshold for the final $K_s$ timesteps:
\begin{equation}
q_{\text{succ}} :=
\mathbf{1}\!\left[
z_k - z_1 \ge \tau_z, \ \forall k \in \{K-K_s+1,\dots,K\}
\right],
\end{equation}
where $\tau_z$ is the lift threshold and $K_s$ is the terminal stability window.

In our experiments, we use:
\begin{itemize}
    \item Grasping: $\tau_z = 0.10$~m, $K_s = 5$,
    \item Scooping: $\tau_z = 0.08$~m, $K_s = 3$.
\end{itemize}

\paragraph{Pushing}
Let $l_k$ denote the Euclidean distance between the object and the push subgoal at timestep $k$. The progress metric is defined as the net reduction in distance:
\begin{equation}
q_{\text{prog}} = \Delta l := l_1 - l_K.
\end{equation}

The success indicator requires the object to remain within a target radius for the final $K_s$ timesteps:
\begin{equation}
q_{\text{succ}} :=
\mathbf{1}\left[
l_k \le \tau_l, \ \forall k \in \{K-K_s+1,\dots,K\}
\right],
\end{equation}
where $\tau_l$ is the distance threshold. In our experiments, we use $\tau_l=0.01$~m, $K_s = 5$.

We use task-specific weights in Eq.~\eqref{eq:single_objective} to balance task performance and object integrity. Table~\ref{tab:design_weights} summarizes the values used in all experiments. $\lambda^{2.5}_{\mathrm{max}}$ and $\lambda^{0}_{\mathrm{max}}$ refer to the weights for 2.5\% and 0\% percentile stress, respectively.

\begin{table}[h]
\centering
\caption{Design objective weights used for different tasks.}
\label{tab:design_weights}
\begin{tabular}{lccccc}
\toprule
Task &
$\lambda_{\text{prog}}$ &
$\lambda_{\text{succ}}$ &
$\lambda^{2.5}_{\mathrm{max}}$ &
$\lambda^{0}_{\mathrm{max}}$ &
$\lambda_{\bar{\sigma}}$ \\
\midrule
Grasping & $1\times10^{2}$ & $2\times10^{2}$ & $2\times10^{-3}$ & $8\times10^{-4}$ & $1\times10^{-2}$ \\
Pushing  & $3\times10^{2}$ & $5\times10^{2}$ & $2\times10^{-3}$ & $8\times10^{-4}$ & $1\times10^{-2}$ \\
Scooping & $1\times10^{2}$ & $2\times10^{2}$ & $2\times10^{-3}$ & $8\times10^{-4}$ & $1\times10^{-2}$ \\
\bottomrule
\end{tabular}
\end{table}

The final design score $J(d)$ is computed by averaging the per-environment scores across $E$ parallel simulation environments.
To prevent degenerate solutions that avoid object interaction and achieve artificially low stress values, stress terms are only included in the average for environments where the task is successful (i.e., $q_{\text{succ}} = 1$). This ensures that low-stress designs are rewarded only when they also accomplish the task.

\subsection{Related Work on Robot Co-Design}
\label{app:related-work}
As discussed in the main paper, most prior co-design approaches operate in relatively low-dimensional design spaces that are amenable to standard derivative-free optimization methods, typically with dimensionality below 15, as summarized in Table~\ref{tab:codesign_comparison}.

\bibliographystyle{plainnat}
\bibliography{references}

\end{document}